\documentclass[runningheads]{llncs}
\usepackage[T1]{fontenc}
\usepackage{graphicx}
\usepackage{subcaption}
\graphicspath{{graphics/}}
\usepackage[numbers,sort&compress]{natbib}
\usepackage{booktabs}
\usepackage[table,xcdraw]{xcolor}
\usepackage[english]{babel}
\usepackage{blindtext}

\begin{document}
\title{A Comprehensive Study of Modern Architectures and Regularization Approaches on CheXpert5000}

\titlerunning{CheXpert5000}
%
\author{Sontje Ihler\inst{1}\orcidID{0000-0002-7495-3141} \and
Felix Kuhnke\inst{2}\orcidID{0000-0002-3003-6755} \and
Svenja Spindeldreier\inst{1}\orcidID{0000-0002-2473-0759}}
\authorrunning{S. Ihler et al.}
%
\institute{Leibniz Universität Hannover, Institute of Mechatronic Systems \and
Leibniz Universität Hannover, Institute for Information Processing
\email{sontje.ihler@imes.uni-hannover.de}
}
\maketitle              
%
\begin{abstract}
Computer aided diagnosis (CAD) has gained an increased amount of attention in the general research community over the last years as an example of a typical limited data application - with experiments on labeled 100k-200k datasets.
Although these datasets are still small compared to natural image datasets like ImageNet1k, ImageNet21k and JFT, they are large for annotated medical datasets, where 1k-10k labeled samples are much more common.
There is no baseline on which methods to build on in the low data regime.
In this work we bridge this gap by providing an extensive study on medical image classification with limited annotations (5k).
We present a study of modern architectures applied to a fixed low data regime of 5000 images on the CheXpert dataset. 
Conclusively we find that models pretrained on ImageNet21k achieve a higher AUC and larger models require less training steps. 
All models are quite well calibrated even though we only fine-tuned on 5000 training samples.
All 'modern' architectures have higher AUC than ResNet50. 
Regularization of Big Transfer Models with MixUp or Mean Teacher improves calibration, MixUp also improves accuracy. 
Vision Transformer achieve comparable or on par results to Big Transfer Models.

\end{abstract}

\begin{figure}
    \centering
    \includegraphics[width=0.49\textwidth]{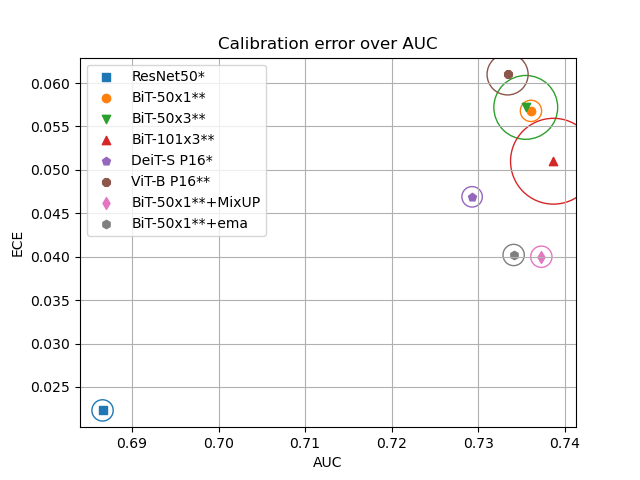}
    \includegraphics[width=0.49\textwidth]{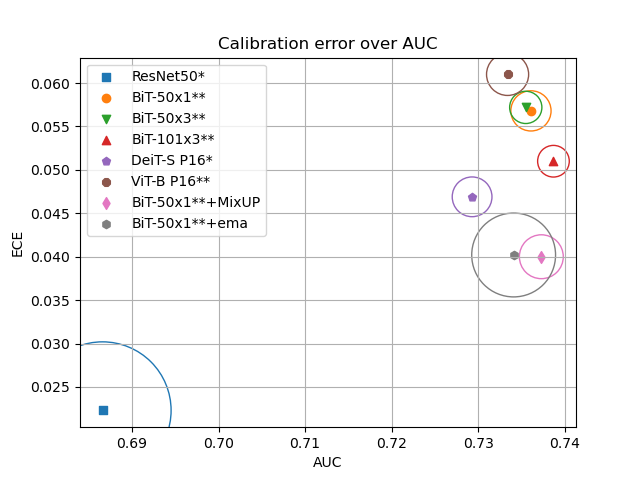}
    \caption{Uncalibrated ECE vs. AUC on CheXpert5000. Metrics are computed on resplit test set. (left) circle size corresponds to number of model parameters, (right) circle size corresponds to our required training steps until convergence. Models pretrained on ImageNet21k achieve a higher AUC. All 'modern' architectures have higher AUC than ResNet50. Larger Models require less training steps. All models are quite well calibrated even though we only fine-tuned on 5000 training samples.  Regularization with MixUp and Mean Teacher (ema) reduces calibration error, MixUp simultaneously increases accuracy for BiT-50x1. \small{*) pretrained on ImageNet **)pretrained on ImageNet21k}}
    \label{fig:teaser}
\end{figure}

\section{Introduction}
Automated analysis from radiology images like X-ray, CT, MRI, and ultrasound is becoming a valuable support tool for diagnosing and treating injuries and diseases. 
With the success of deep learning for image classification more and more applications come within reach of becoming standard tools for radiologists.
The tremendous increase in research on medical image classification is usually fueled by strategies from the natural image domain (computer vision). 
At the same time computer aided diagnosis (CAD) has gained an increased amount of attention in the general research community over the last years as an example of a typical limited data application - with experiments on labeled 100k-200k datasets.
A common approach to handle limited data regimes is transfer learning. 
Models are pretrained on large natural image datasets and then afterwards fine-tuned to the medical application. 
However, radiology images are quite different from natural (photographic) images and computer vision (CV) methods need to be adapted to the medical image domain. 
To complicate matters, large labeled datasets are widely available in the CV community and associated methods build on this fact.
In the medical domain, annotations are much harder to obtain. 
These burdens are a major bottleneck in the development and transfer of CV model to image classification applications in radiology.
These connections also generate a phenomenon in the medical image community.
While in CV new methods are usually compared in large studies and promoted as general problem solvers, in the medical community specific adaptations of a solution are presented. 
For newcomers, an overview of the available baselines is often missing.

In the recent past, great improvements have been achieved in transfer learning on natural images. 
An elementary component for successful transfer learning is the scaling of model capacity and pretraining datasets. 
Bigger is better. 
So-called Big Transfer Models (BiT)~\cite{kolesnikov2020big} have reached the state of the art for many visual benchmarks in the natural image domain. 
BiT models are tweaked ResNet~\cite{he2016identity} variants which are optimized for transfer performance.
Recently,~\cite{mustafa2021supervised} applied these models to the CheXpert dataset~\cite{irvin2019chexpert} (among others) to study the transfer effect to the medical image domain. 
The authors found these models to outperform a ResNet50 baseline on all studied datasets. 
The overall findings in this work are very promising, but it should be noted that CheXpert is comparably large for a medical dataset with over 200k X-ray images. 
Is it not uncommon for medical dataset to be drastically smaller (four-digit range).
The authors do indeed study the effect of fewer labels, by reducing the amount of transfer labels and comparing to the full ResNet50 baseline. 
This top-down approach naturally provides information on robustness to data reduction, but it still misses a very important point. 
For safety-critical applications, the ultimate goal should always be to achieve the greatest accuracy on the greatest amount of data available.
\textbf{An important question is therefore, if only few examples are available, how can we get the most out of them}.
This just mentioned top-down approach has limitations for directly comparing methods specifically in a low data regime. 
We believe that a direct comparison of methods is essential to improve performance on low medical data regimes. 
One might argue that performance on small data regimes will become obsolete in the near future due to federated learning~\cite{rieke2020future} but we believe that large datasets will still be correlated to economic interest. 
High quality labels from specialized domain experts are expensive and are subject to chance/privilege for individual patients (like selection of relevant diseases and patient groups).
We therefore provide a comprehensive study of modern architectures and regularization approaches at a fixed low data regime to fill the gap of method comparisons to improve performance specifically on limited medical data. We not only test Big Transfer models in a very small data regime but also Vision Transformers~\cite{dosovitskiy2020image}, another model class pretrained on huge datasets which have recently outperformed CNNs. 

Our contribution is a study of these modern architectures/methods applied to a fixed low data regime of 5000 images on the CheXpert dataset, which to our knowledge does not exist in this form. We include:
\begin{enumerate}
\item Big Transfer models~\cite{kolesnikov2020big} at varying sizes,
\item ConvNets vs. Vision Transformer~\cite{dosovitskiy2020image,touvron2021training},
\item gains of standard regularization: Mean Teacher~\cite{tarvainen2017mean} and MixUp~\cite{zhang2017mixup}, 
\item public data splits to enable direct method comparison in the future.
\end{enumerate}

\section{Related Work}
This work is mainly inspired by 'Supervised Transfer Learning at Scale for Medical Imaging'~\cite{mustafa2021supervised}. This work compares accuracy on the CheXpert dataset for Big Transfer (BiT) models~\cite{kolesnikov2020big} of different sizes pretrained on datasets of different sizes.  
BiT models are updated/improved (wide) ResNet-v2 models~\cite{he2016identity} which are optimized for transfer learning.
Batch normalization is replaced with group normalization. 
Studied BiT models are BiT-50x1, BiT-101x1, BiT-50x3, and BiT-101x3, the baseline is a standard Resnet50~\cite{he2016deep}. 
The BiT models are pretrained on three datasets of increasing size: 1. \emph{ImageNet}~\cite{russakovsky2015imagenet}: 1.3M image, 1000 classes, multi-class labels, 2. \emph{ImageNet-21k }~\cite{deng2009imagenet}: 14M images, 21000 classes, multi-class labels and 3. non-public \emph{JFT-300M}~\cite{sun2017revisiting}: 300M images, 18k classes, average of 1.29 labels per image, with pretrained weights for the former two publicly available.
In~\cite{mustafa2021supervised}, all BiT models outperform the ResNet50 baseline (all pretrained on ImageNet) on all medical datasets.

Generally accuracy increases for increase in model size and increased number of pretraining samples. 
There are more improvements from ImageNet to ImageNet21k pretraining than from ImageNet21k to JFT. 
Larger models are more data efficient and
models pretrained on larger datasets are more data efficient. A BiT-50x1 model pretrained on ImageNet only requires 75\% of the data to achieve the same accuracy as a standard Resnet50. 
A BiT-101x3 model pretrained on Imagenet21K~\cite{deng2009imagenet} only requires 49\%. 
They find that larger models converge faster (taking only about half of the time, sometimes even less) and larger models show better generalization performance to other datasets in the same domain (small domain shift).
The authors find improved calibration for the larger models on a dermatology dataset~\cite{liu2020deep} (derived from camera images which are presumably closer to natural images than X-ray imaging) but not for the CheXpert dataset.

For applications in safety-critical applications like medical image analysis it is essential that models do not only provide highly accurate predictions on the test domain (as well as under domain shift) but they should also be well calibrated. 
For a well predicted model, the predicted numerical output (referred to as confidence) should correlate to the real accuracy of the model. 
The predicted confidence then provides a reliable measure of certainty/uncertainty of a prediction. In the last years the best performing convnet models like ResNet, Dense-Net and Efficient-Net unfortunately showed poor calibration properties~\cite{guo2017calibration}. 
Recently Minderer et al.~\cite{minderer2021revisiting} have shown that newer model architectures MLP-Mixer~\cite{tolstikhin2021mlp} and Vision Transformer (ViT)~\cite{dosovitskiy2020image} not only achieve higher accuracy on ImageNet but also improved calibration. This also applies to BiT models to some extent. These findings apply also to performance under distribution shift. These findings have been established in natural images at large scale. It is not clear how these results transfer to very limited medical data. 

ViT models been successfully applied to medical images with results on par with a ResNet50 baseline~\cite{matsoukas2021time}. 
We hope to transfer their successes to our case as well.

\section{Experiments}
We study BiT and ViT models in our experiments, as well as established regularization methods Mean Teacher and MixUp. For comparison to older works we compare all results to ResNet50, a very common backbone in medical image classification. 
All experiments are based on pretrained models which are then fine-tuned to CheXpert5000. Our baseline ResNet50 and DeiT are pretrained on ImageNet, while all other models are pretrained on Imagenet21k. We use models and pretrained weights from the timm library~\cite{rw2019timm} (PyTorch). For all experiments we use the weak image augmentation described in \ref{dataset}. The best model has highest AUC on the validation set. We trained all models on five different train sets with 5000 images each and provide mean and standard deviation of the five best models. All 5k models are trained on the same five train sets. For comparison we also trained ResNet50 on the full CheXpert dataset.

\subsection{CheXpert5000}\label{dataset}

We perform our method study on CheXpert~\cite{irvin2019chexpert}, a publicly available dataset of chest X-ray images. It contains 224,316 radiographs from 65,240 patients. The validation set was created from 234 manually annotated X-rays. The train set was automatically extracted from patients' reports. The labels on the train set were created using natural language processing on these reports. To handle incomplete diagnosis in the patients' reports there are 4 categories for 14 observations (diseases). For each observation the categories are (1) certain disease observation, (u) uncertain disease observation, (-) disease not mentioned, and (0) disease ruled out. The dataset also provides age, gender and view point. Most images are frontal view and PA.

\textbf{Preprocessing of CheXpert}:
Mustafa et al.~\cite{mustafa2021supervised} have found that comparing models on the provided validation set is unreliable due to it's small size (234 images).
We follow their protocol and resplit the original train set into a new train (75\%: 124,664 images), validation (10\%: 16,989 images) and test set (15\%: 25,205) based on patient id. 

We further follow the common practice to reduce the number of classes to 5 (multi-label classification)and map uncertain labels (u) to 1 and missing observations (-) to 0.
To study the effect of limited annotations we create five subsets of the train set of 5000 labeled samples each which are all frontal view and PA, half male, half female.
We therefore also reduce the validation and test set to frontal view and PA (validation set: 12,115 images and test set: 18,363 images). We provide lists of all mentioned splits and subsets\footnote{https://gitlab.uni-hannover.de/sontje.ihler/chexpert5000}. All subsets are created by random sampling (uniform distribution) so that all datasets show the same label distribution as the original dataset.

Mustafa et al. studied varying image resolution as model input and found no improvement in accuracy on the CheXpert dataset with resolution higher than 224x224. We therefore use the 'small' version of the dataset with image size 320x320 which is a lot more accessible due to its smaller memory consumption. All images are normalized with mean 0.5 and standard deviation 0.5.  

Large validation sets are obviously not representative for a true small data regime, however the focus of this work is an analysis of the potential of the varying model architectures which requires a robust stop criterion. The effect of small validation sets and resulting fluctuation in convergence of the validation loss are subject of future work.

\textbf{Image augmentations} in the medical image domain is a risky task, we therefore only employ limited augmentations again following~\cite{mustafa2021supervised}. Spatial Transformation: We scale images to 248x248, perform random rotation by angle $\alpha\sim Uniform(-20, 20)$, random crops to 224x224, random horizontal flips with probability 50\%. Chromatic Transformation: We employ random brightness and contrast jitter with factor 0.2.

\subsection{Finetuning: Models and Augmentation}
Our \textbf{ResNet50 baselines} were fine-tuned using SGD with learning rate 0.003, momentum 0.9, weight decay 2e-5, 10 warm up epochs, and plateau scheduler (patience: 10 epochs, decay rate : 0.1). Training was terminated when the learning rate dropped below 1e-6.  We used batchsize 512~\cite{kolesnikov2020big} and batch size 32 which we found to improve results.

We study the following \textbf{BiT models}~\cite{kolesnikov2020big}: BiT-50x1, BiT-50x3 and BiT-101x3 which can also be found in~\cite{mustafa2021supervised}.
For fine-tuning our BiT models we follow the BiT-HyperRule which proposes SGD with an initial learning rate of 0.003, momentum 0.9, and batch size 512. During fine-tuning the learning rate should be decayed by a factor of 10 at 30\%, 60\% and 90\% of the training steps. For datasets with less than 20k samples, the authors propose 500 fine-tuning steps with a batch size of 512 and no MixUp~\cite{kolesnikov2020big}. We therefore planned on training all 5k models for 60 epochs but we found that BiT-50x1 did not universally converge in that time. We therefore trained BiT-50x1 for 100 epochs ($\approx$ 900 steps). We also found that finetuning the BiT models with a smaller batch size yielded better results. We therefore also fine-tuned all BiT models with our baseline protocol with a batch size of 32.

We employ two configurations of \textbf{ViT models}: ViT-B and DeiT. (ViT-B) In accordance to the finding of~\cite{steiner2021train} that more pretraining data generally outperforms any data augmentation or regularization, we use models pretrained on ImageNet21k . We therefore do not use any regularization or advanced data augmentation (apart from the augmentation described in section \ref{dataset}). Following their fine-tuning protocol for their smalles datasets, we use SGD with a momentum of 0.9, use a learning rate of 0.003 and a batchsize of 512. We train for 500 steps with cosine decay learning rate and 3 warm-up epochs, as well as gradient clipping at norm 1. (DeiT-S) For the DeiT model we follow the training protocol of~\cite{matsoukas2021time}. We use DeiT-S with a patch size 16. We use Adam optimizer with learning rate 1e-4, momentum 0.9, and weight decay 1e-5, and a plateau scheduler with minimal learning rate 1e-6 (warm up for 10 epochs). Batch size is 64.

\textbf{Mean Teacher and MixUp:} Low data regime strategies like contrastive and consistency learning are based on (strong) image augmentations. This is always challenging for medical image diagnosis as perturbations can alter images in a way that eliminate disease relevant landmarks and features. We therefore study the effect of two established low data regime approaches Mean Teacher and MixUp that do not rely on strong data augmentation. 
We apply both regularizations to BiT-50x1. We follow our baseline protocol with batch size 32 and set $\alpha$ to 0.5 for MixUp regularization with a MixUp probability of 50\%.

\begin{table}[t]
\centering
\caption{BiT models vs. ResNet50 on 5k train set. The BiT models are trained according to the BiT-HyperRule. We used the same batch size for ResNet50. We provide the amount of images the model has seen during training (image iter.) before convergence of validation loss. To enable a comparison on a larger scale we also provide results from training BiT-50x1 on the full train set.}
\label{tab:bit-bs512}
\begin{tabular}{@{}ccccccc@{}}
\toprule
\textbf{Model} &
  \textbf{params.} &
  \textbf{pretr.} &
  \textbf{\begin{tabular}[c]{@{}c@{}}training\\ samples\end{tabular}} &
  \textbf{\begin{tabular}[c]{@{}c@{}}batch\\ size\end{tabular}} &
  \textbf{\begin{tabular}[c]{@{}c@{}}image\\ iter.\end{tabular}} &
  \textbf{\begin{tabular}[c]{@{}c@{}}AUC\\ (resplit)\end{tabular}} \\ \midrule
\textbf{ResNet50}  & 23.51M  & \textbf{in1k} & 5000  & 512 & 438k & 0.5615\textpm.0028\\ \midrule
\textbf{BiT-50x1}  & 23.51M  & in21k         & 5000  & 512 & 256k   & 0.7206\textpm.0021 \\
\textbf{BiT-50x3}  & 211.19M & in21k         & 5000  & 512 & 112k   & 0.7218\textpm.0009 \\
\textbf{BiT-101x3} & 381.81M & in21k         & 5000  & 512 & 99k    & \textbf{0.7317\textpm.0015} \\ \midrule
\textbf{BiT-50x1}  & 23.51M  & in21k         & 89944* & 512 & 2330k  & 0.7718          \\
\textbf{BiT-50x1}  & 23.51M  & in21k         & 124663**   & 512 & 2070k  & 0.7714          \\ \bottomrule
\multicolumn{7}{l}{*) all Frontal/AP views from full resplit train set **) full resplit train set}
\end{tabular}%

\end{table}
\begin{table}[t]
\centering
\caption{BiT models vs. ResNet50 on 5k train set trained according to baseline protocol with a batch size of 32. Reducing the batch size (in comparison to BiT-HyperRule) decreases training time drastically while also improving the accuracy for all models.}
\label{tab:bit-bs32}
\begin{tabular}{lrrrrcc}
\hline
\multicolumn{1}{c}{\textbf{Model}} &
  \multicolumn{1}{c}{\textbf{params.}} &
  \multicolumn{1}{c}{\textbf{pretr.}} &
  \multicolumn{1}{c}{\textbf{\begin{tabular}[c]{@{}c@{}}batch\\ size\end{tabular}}} &
  \multicolumn{1}{c}{\textbf{\begin{tabular}[c]{@{}c@{}}image\\ iter.\end{tabular}}} &
  \textbf{\begin{tabular}[c]{@{}c@{}}AUC\\ (resplit)\end{tabular}} &
  \textbf{\begin{tabular}[c]{@{}c@{}}AUC\\ (official)\end{tabular}} \\ \hline
\textbf{ResNet50}  & 23.51M  & \textbf{in1k} & 32 & 487 & 0.6866\textpm.0030 &   0.7949\textpm.0106\\ \hline
\textbf{BiT-50x1}  & 23.51M  & in21k & 32 & 42k          & 0.7361\textpm.0018 & \textbf{0.8380\textpm.0095}\\
\textbf{BiT-50x3}  & 211.19M & in21k & 32 & 27k          & 0.7355\textpm.0029          & 0.8359\textpm.0169\\
\textbf{BiT-101x3} & 381.81M & in21k & 32 & \textbf{26k} & \textbf{0.7388\textpm.0008} & 0.8342\textpm+.0212 \\ \hline
\end{tabular}%
\end{table}

\section{Results}\label{sec:results}
To evaluate the prediction quality of all models fine-tuned models we provide the common metric AUC (also: AUROC), i.\,e. area under the curve for the ROC curve (true positive rate vs. false positive rate), and additionally we also provide class-wise AUPRC, i.\,e. the area under the PR curve (precision vs. recall) in the appendix. To quantify the calibration we compute ECE i.\,e. the expected calibration error. The calibration error is computed based on histogram binning with equal number of samples per bin (31 bins). We only compute the calibration on our test split, as the official validation set is too small to obtain representative results.

\begin{table}[t]
\centering
\caption{ResNet50 variations vs. Vision Transformer. The ViT-B P16* model achieves comparable results to Bit-50x1. }
\label{tab:transformer}
\begin{tabular}{lcccccc}
\hline
\textbf{Model} &
  \textbf{params.} &
  \textbf{pretr.} &
  \textbf{\begin{tabular}[c]{@{}c@{}}batch\\ size\end{tabular}} &
  \textbf{\begin{tabular}[c]{@{}c@{}}image\\ iter.\end{tabular}} &
  \textbf{\begin{tabular}[c]{@{}c@{}}AUC\\ (resplit)\end{tabular}} &
  \textbf{\begin{tabular}[c]{@{}c@{}}AUC\\ (official)\end{tabular}} \\ \hline
\textbf{ResNet50}  & 23.51M  & \textbf{in1k} & 32 & 487 & 0.6866\textpm.0030 &   0.7949\textpm.0106\\ 
\textbf{BiT-50x1}   & 23.51M & in21k         & 32 & 42k & \textbf{0.7361\textpm.0018} & \textbf{0.8380\textpm.0095}\\ \hline
\textbf{DeiT-S P16} & 21.67M  & in1k    & 64  & 41k   & 0.7293\textpm.0035  & 0.8161\textpm.0173\\
\textbf{ViT-B P16}  & 87.46M  & in21k   & 512 & 352k  & 0.6394\textpm.0070  & 0.7704\textpm.0054\\ 
\textbf{ViT-B P16*}  & 87.46M  & in21k  & 64  & 46k   & 0.7334\textpm.0032  & 0.8299\textpm.0139 \\ \hline
\multicolumn{5}{l}{*)ViT-B trained on DeiT-S fine-tuning protocol} 
\end{tabular}%

\end{table}

\begin{table}[t]
\centering
\caption{Vanilla BiT-50x1 (none) vs. regularized BiT-50x1 using MixUp and Mean Teacher. We provide results for the teacher (ema) model as the teacher generally achieves higher accuracies. }
\label{tab:regularization}
\begin{tabular}{lcccccc}
\hline
\textbf{Regularization} &
  \textbf{params} &
  \textbf{pretr.} &
  \textbf{\begin{tabular}[c]{@{}c@{}}batch\\ size\end{tabular}} &
  \textbf{\begin{tabular}[c]{@{}c@{}}image\\ iter.\end{tabular}} &
  \textbf{\begin{tabular}[c]{@{}c@{}}AUC\\ (resplit)\end{tabular}} &
  \textbf{\begin{tabular}[c]{@{}c@{}}AUC\\ (official)\end{tabular}} \\ \hline
\textbf{none}         & 23.51M & in21k & 32 & 42k  & 0.7361\textpm.0018 & \textbf{0.8380\textpm.0095}\\ 
\textbf{MixUP}        & 23.51M & in21k & 32 & 50k  & 0.\textbf{7373\textpm.0005} & 0.8313\textpm.0116 \\
\textbf{Mean Teacher} & 23.51M & in21k & 32 & 190k & 0.7341\textpm.0141    & 0.8267\textpm.0103\\ \hline
\end{tabular}%
\end{table}

For the highest possible comparisons with other works on CheXpert we provide accuracy measures on our new test split of automatically labeled samples as well as the provided validation set of manual annotations. We provide the numbers of parameters for the used models as well as the amount of images the best models has seen during training until convergence of validation AUC. The computation steps can be computed by dividing image iterations by the batch size.

In Table \ref{tab:bit-bs512} we show BiT models trained according to BiT-Hyperrule. Equivalent to the results on the full CheXpert datset, larger models show higher accuracy.  The BiT-Hyperrule recommends a high batch size of 512 - but we saw great improvement regarding accuracy and training time when reducing the batchsize for CheXpert, see Table~\ref{tab:bit-bs32}. Fine-tuned with a batch size of 32 the BiT models arrive at their optimum in significantly less training steps (less than 28 times) with simultaneously higher accuracy. The BiT models based on ResNet-v2 architecture outperform standard ResNet50 not only on large but also on our very small data regime. ResNet50 and BiT-50x1 are almost identical in architecture and ResNet50 can be interchanged easily by BiT-50x1 in existing frameworks to probably gain a performance boost and shorter training times.

In Table~\ref{tab:transformer} we compare results for ResNet variants ResNet50 and BiT-50x1 (ResNet-v2) to ViT models. Even though ViT models are infamous to require large amounts of data, they show great performance in this very small data regime outperform the ResNet50 baseline. The DeiT model is not yet quite on par with the BiT model, however the bigger ViT model shows comparable performance (at least when trained with the DeiT-S protocol). These findings are similar to~\cite{matsoukas2021time}. 

We finally show the effect of regularization in the small data regime in Table~\ref{tab:regularization}. To our surprise regularization has little effect on accuracy. However, both regularizations improve calibration, see Fig.~\ref{fig:teaser}.

Note: Accuracy on the automatic labels is considerably lower than on the manual labels. This is on par with the works of~\cite{mustafa2021supervised, azizi2021big}. We provide the class-wise AUC und AUPRC in the appendix for all five classes on the official validation set.

\section{Conclusion and Discussion}
In this work we presented a method study of modern architectures applied to a fixed low data regime of 5000 images on the CheXpert dataset and provide subsets and data splits for reproducibility. Conclusively we find/verify that model pretrained on ImageNet21k achieve a higher AUC and larger models require less training steps. All models are quite well calibrated even though we only fine-tuned on 5000 training samples. All 'modern' architectures have higher AUC than ResNet50. Regularization of BiT-50x1 with MixUp or Mean Teacher improves calibration and accuracy. Vision Transformer achieve comparable or on par results to BiT-50x1.
While BiT-50x1 is one of many updates of the ResNet variants, ViTs are still in their infancy for small data regime szenarios. As ViTs can outperform CNNs in a large data regime, it may therefore only be a question of time until they outgrow CNNs for low data regimes.

\bibliographystyle{splncs04}
\bibliography{bibliography}

\begin{thebibliography}{10}
\providecommand{\url}[1]{\texttt{#1}}
\providecommand{\urlprefix}{URL }
\providecommand{\doi}[1]{https://doi.org/#1}

\bibitem{azizi2021big}
Azizi, S., Mustafa, B., Ryan, F., Beaver, Z., Freyberg, J., Deaton, J., Loh,
  A., Karthikesalingam, A., Kornblith, S., Chen, T., et~al.: Big
  self-supervised models advance medical image classification. In: Proceedings
  of the IEEE/CVF International Conference on Computer Vision. pp. 3478--3488
  (2021)

\bibitem{deng2009imagenet}
Deng, J., Dong, W., Socher, R., Li, L.J., Li, K., Fei-Fei, L.: Imagenet: A
  large-scale hierarchical image database. In: 2009 IEEE conference on computer
  vision and pattern recognition. pp. 248--255. Ieee (2009)

\bibitem{dosovitskiy2020image}
Dosovitskiy, A., Beyer, L., Kolesnikov, A., Weissenborn, D., Zhai, X.,
  Unterthiner, T., Dehghani, M., Minderer, M., Heigold, G., Gelly, S., et~al.:
  An image is worth 16x16 words: Transformers for image recognition at scale.
  arXiv preprint arXiv:2010.11929  (2020)

\bibitem{guo2017calibration}
Guo, C., Pleiss, G., Sun, Y., Weinberger, K.Q.: On calibration of modern neural
  networks. In: International Conference on Machine Learning. pp. 1321--1330.
  PMLR (2017)

\bibitem{he2016deep}
He, K., Zhang, X., Ren, S., Sun, J.: Deep residual learning for image
  recognition. In: Proceedings of the IEEE conference on computer vision and
  pattern recognition. pp. 770--778 (2016)

\bibitem{he2016identity}
He, K., Zhang, X., Ren, S., Sun, J.: Identity mappings in deep residual
  networks. In: European conference on computer vision. pp. 630--645. Springer
  (2016)

\bibitem{irvin2019chexpert}
Irvin, J., Rajpurkar, P., Ko, M., Yu, Y., Ciurea-Ilcus, S., Chute, C.,
  Marklund, H., Haghgoo, B., Ball, R., Shpanskaya, K., et~al.: Chexpert: A
  large chest radiograph dataset with uncertainty labels and expert comparison.
  In: Proceedings of the AAAI conference on artificial intelligence. vol.~33,
  pp. 590--597 (2019)

\bibitem{kolesnikov2020big}
Kolesnikov, A., Beyer, L., Zhai, X., Puigcerver, J., Yung, J., Gelly, S.,
  Houlsby, N.: Big transfer (bit): General visual representation learning. In:
  European conference on computer vision. pp. 491--507. Springer (2020)

\bibitem{liu2020deep}
Liu, Y., Jain, A., Eng, C., Way, D.H., Lee, K., Bui, P., Kanada, K.,
  de~Oliveira~Marinho, G., Gallegos, J., Gabriele, S., et~al.: A deep learning
  system for differential diagnosis of skin diseases. Nature medicine
  \textbf{26}(6),  900--908 (2020)

\bibitem{matsoukas2021time}
Matsoukas, C., Haslum, J.F., S{\"o}derberg, M., Smith, K.: Is it time to
  replace cnns with transformers for medical images? arXiv preprint
  arXiv:2108.09038  (2021)

\bibitem{minderer2021revisiting}
Minderer, M., Djolonga, J., Romijnders, R., Hubis, F., Zhai, X., Houlsby, N.,
  Tran, D., Lucic, M.: Revisiting the calibration of modern neural networks.
  Advances in Neural Information Processing Systems  \textbf{34} (2021)

\bibitem{mustafa2021supervised}
Mustafa, B., Loh, A., Freyberg, J., MacWilliams, P., Wilson, M., McKinney,
  S.M., Sieniek, M., Winkens, J., Liu, Y., Bui, P., et~al.: Supervised transfer
  learning at scale for medical imaging. arXiv preprint arXiv:2101.05913
  (2021)

\bibitem{rieke2020future}
Rieke, N., Hancox, J., Li, W., Milletari, F., Roth, H.R., Albarqouni, S.,
  Bakas, S., Galtier, M.N., Landman, B.A., Maier-Hein, K., et~al.: The future
  of digital health with federated learning. NPJ digital medicine
  \textbf{3}(1), ~1--7 (2020)

\bibitem{russakovsky2015imagenet}
Russakovsky, O., Deng, J., Su, H., Krause, J., Satheesh, S., Ma, S., Huang, Z.,
  Karpathy, A., Khosla, A., Bernstein, M., et~al.: Imagenet large scale visual
  recognition challenge. International journal of computer vision
  \textbf{115}(3),  211--252 (2015)

\bibitem{steiner2021train}
Steiner, A., Kolesnikov, A., Zhai, X., Wightman, R., Uszkoreit, J., Beyer, L.:
  How to train your vit? data, augmentation, and regularization in vision
  transformers. arXiv preprint arXiv:2106.10270  (2021)

\bibitem{sun2017revisiting}
Sun, C., Shrivastava, A., Singh, S., Gupta, A.: Revisiting unreasonable
  effectiveness of data in deep learning era. In: Proceedings of the IEEE
  international conference on computer vision. pp. 843--852 (2017)

\bibitem{tarvainen2017mean}
Tarvainen, A., Valpola, H.: Mean teachers are better role models:
  Weight-averaged consistency targets improve semi-supervised deep learning
  results. Advances in neural information processing systems  \textbf{30}
  (2017)

\bibitem{tolstikhin2021mlp}
Tolstikhin, I.O., Houlsby, N., Kolesnikov, A., Beyer, L., Zhai, X.,
  Unterthiner, T., Yung, J., Steiner, A., Keysers, D., Uszkoreit, J., et~al.:
  Mlp-mixer: An all-mlp architecture for vision. Advances in Neural Information
  Processing Systems  \textbf{34} (2021)

\bibitem{touvron2021training}
Touvron, H., Cord, M., Douze, M., Massa, F., Sablayrolles, A., J{\'e}gou, H.:
  Training data-efficient image transformers \& distillation through attention.
  In: International Conference on Machine Learning. pp. 10347--10357. PMLR
  (2021)

\bibitem{rw2019timm}
Wightman, R.: Pytorch image models.
  \url{https://github.com/rwightman/pytorch-image-models} (2019).
  \doi{10.5281/zenodo.4414861}

\bibitem{zhang2017mixup}
Zhang, H., Cisse, M., Dauphin, Y.N., Lopez-Paz, D.: mixup: Beyond empirical
  risk minimization. arXiv preprint arXiv:1710.09412  (2017)

\end{thebibliography}

\newpage
\appendix

\section{Additional Results}

\begin{figure}
    \centering
     \includegraphics[width=0.19\linewidth]{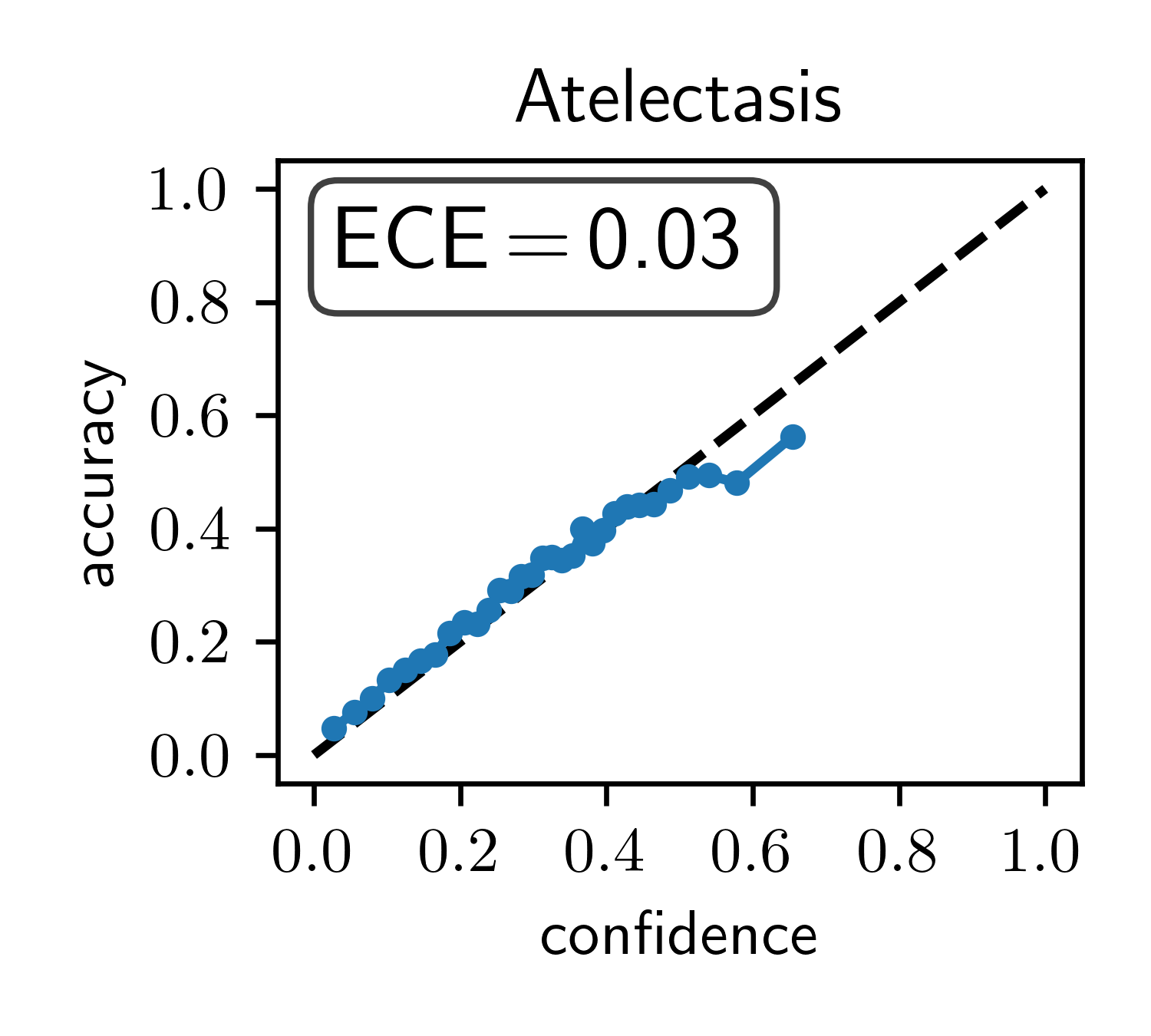}
     \includegraphics[width=0.19\textwidth]{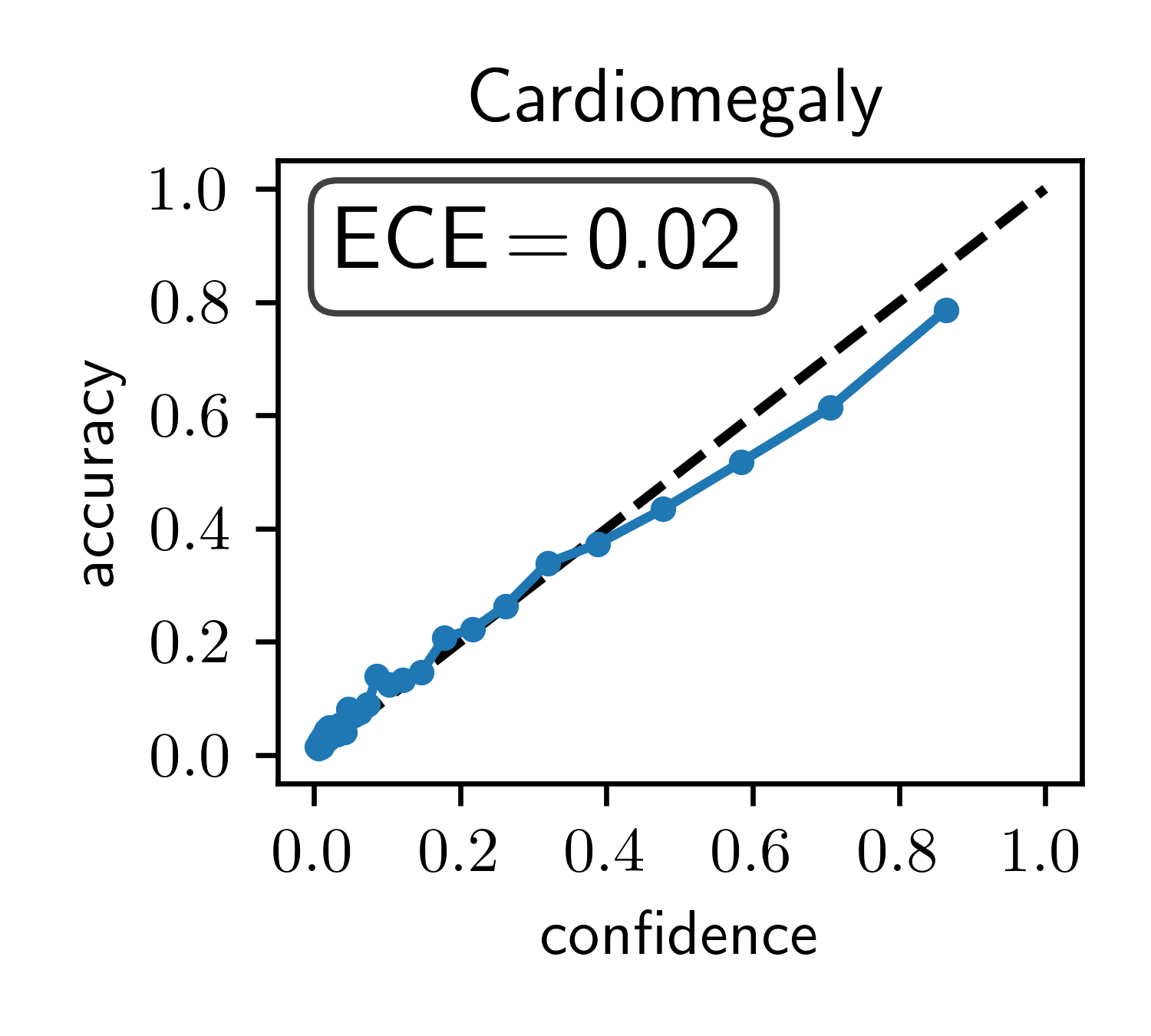}
     \includegraphics[width=0.19\textwidth]{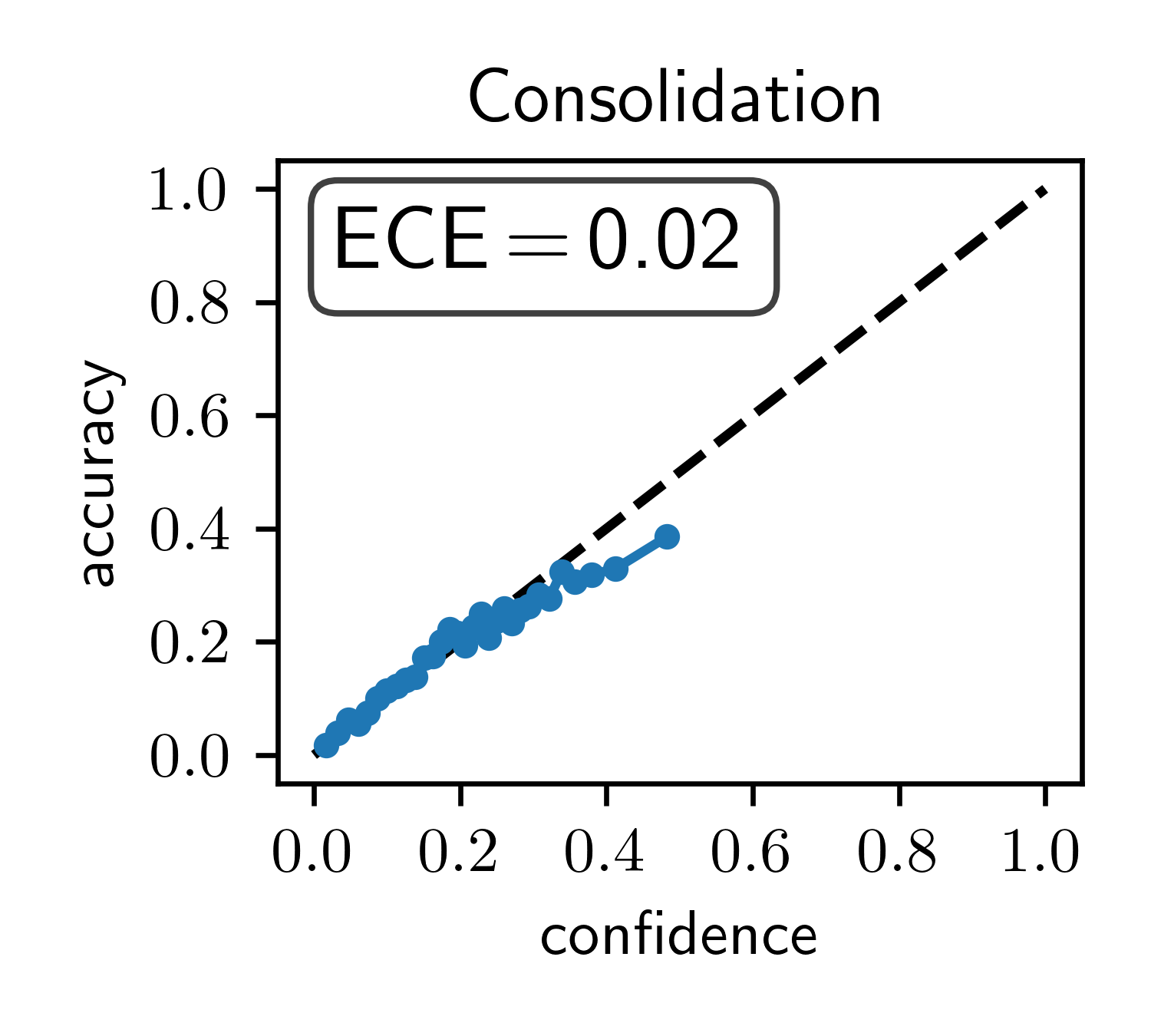}
     \includegraphics[width=0.19\textwidth]{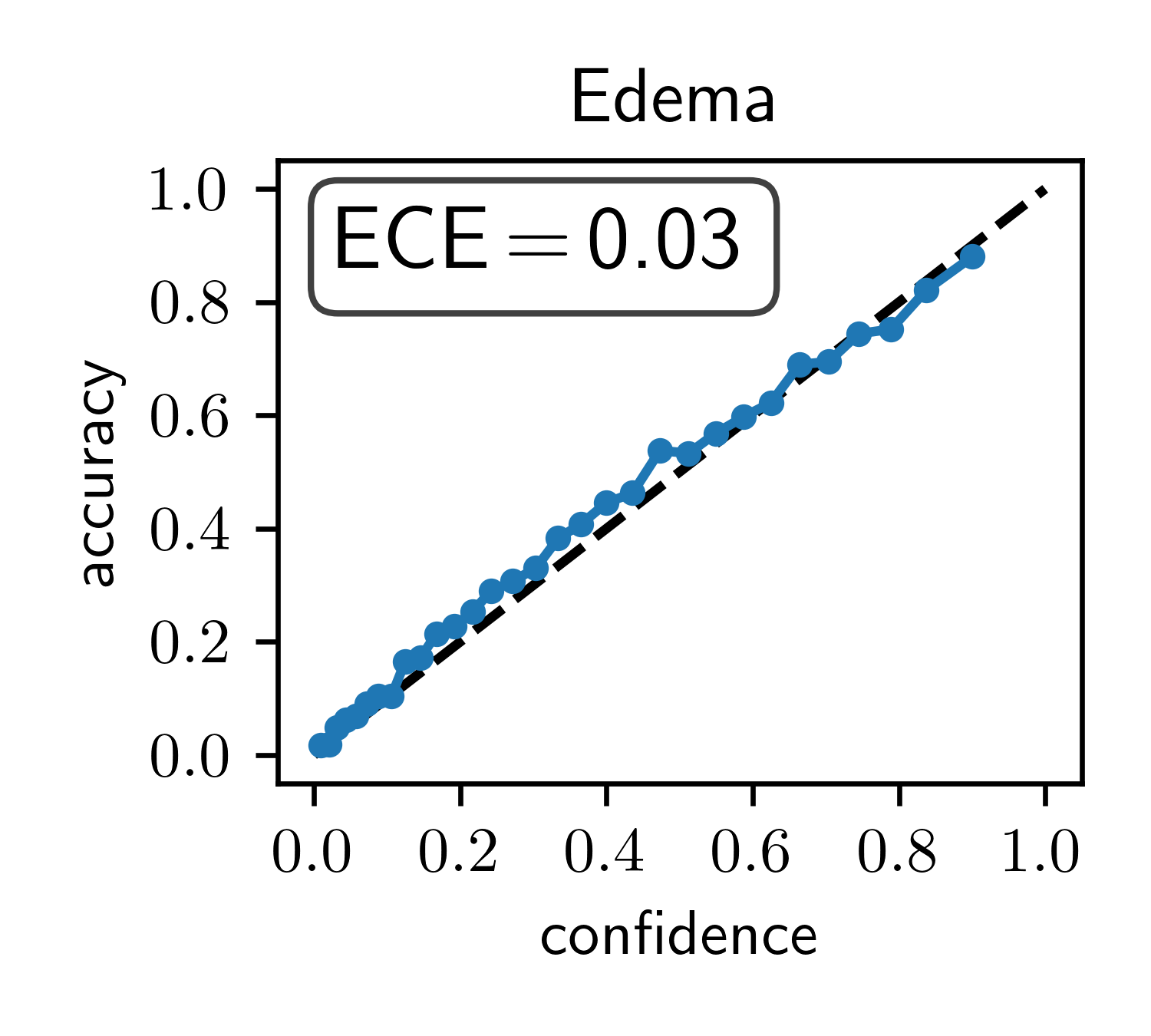}
     \includegraphics[width=0.19\textwidth]{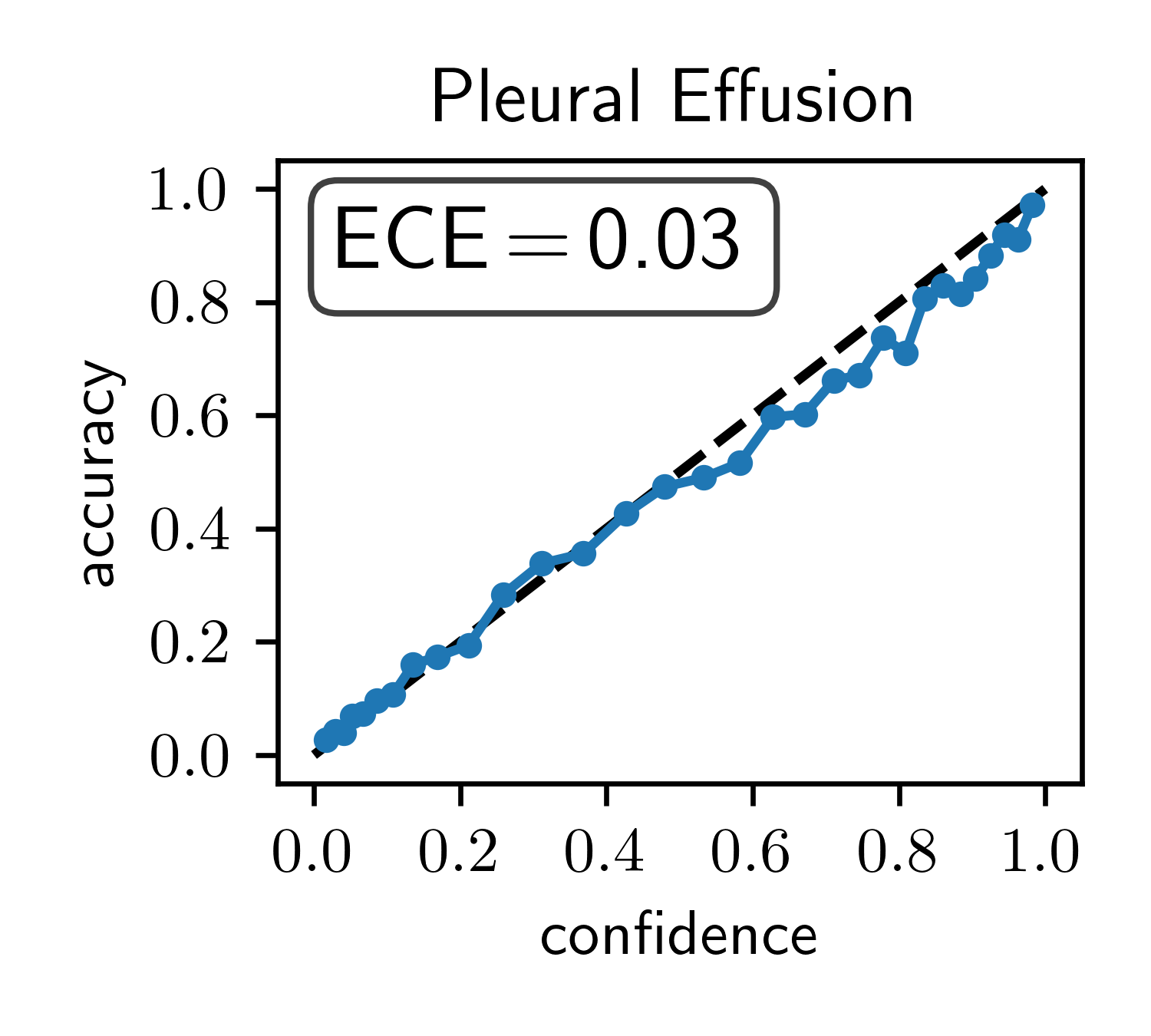}
    \caption*{BiT-50x1 with all frontal/AP samples}
    
     \includegraphics[width=0.19\linewidth]{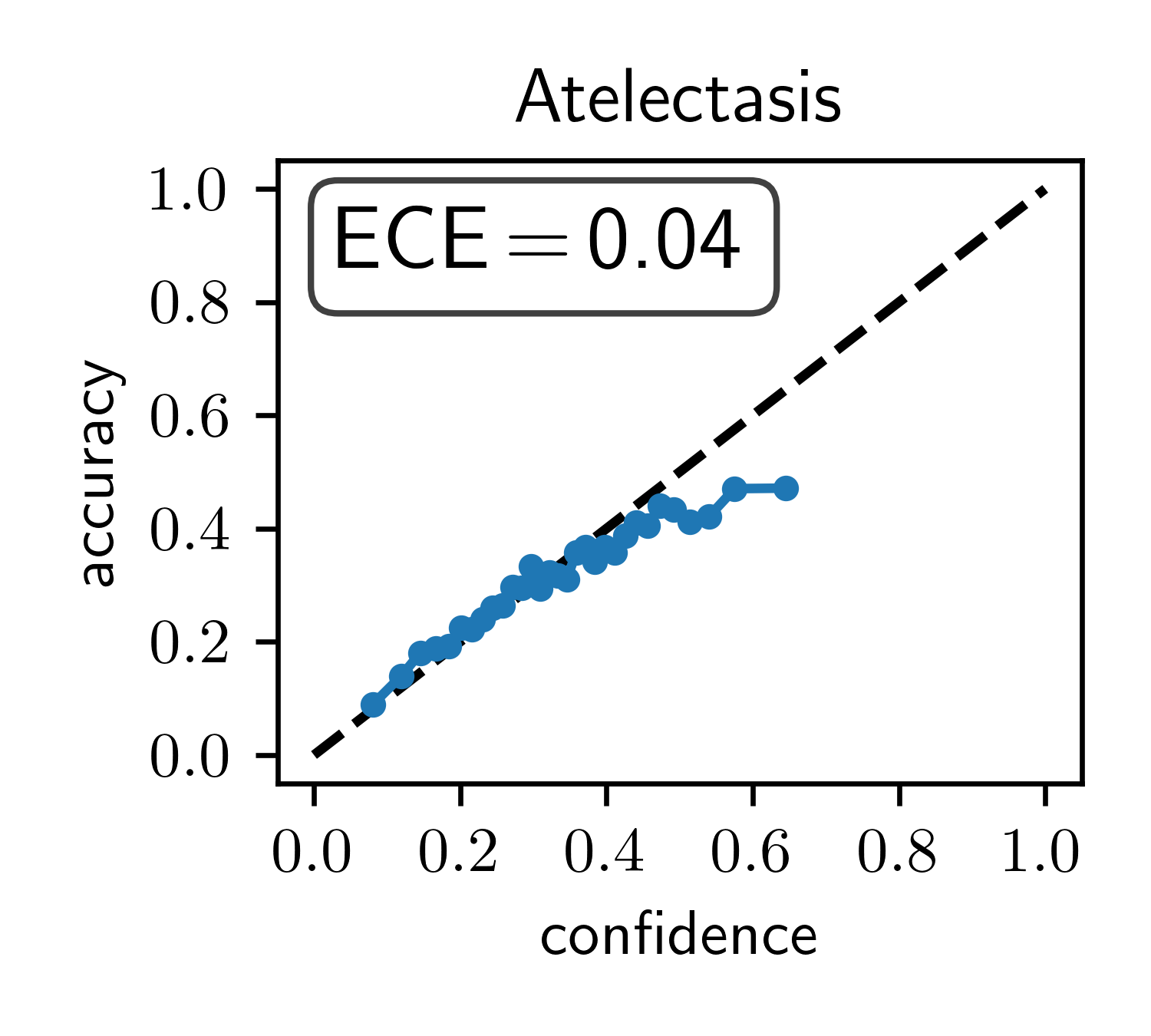}
     \includegraphics[width=0.19\textwidth]{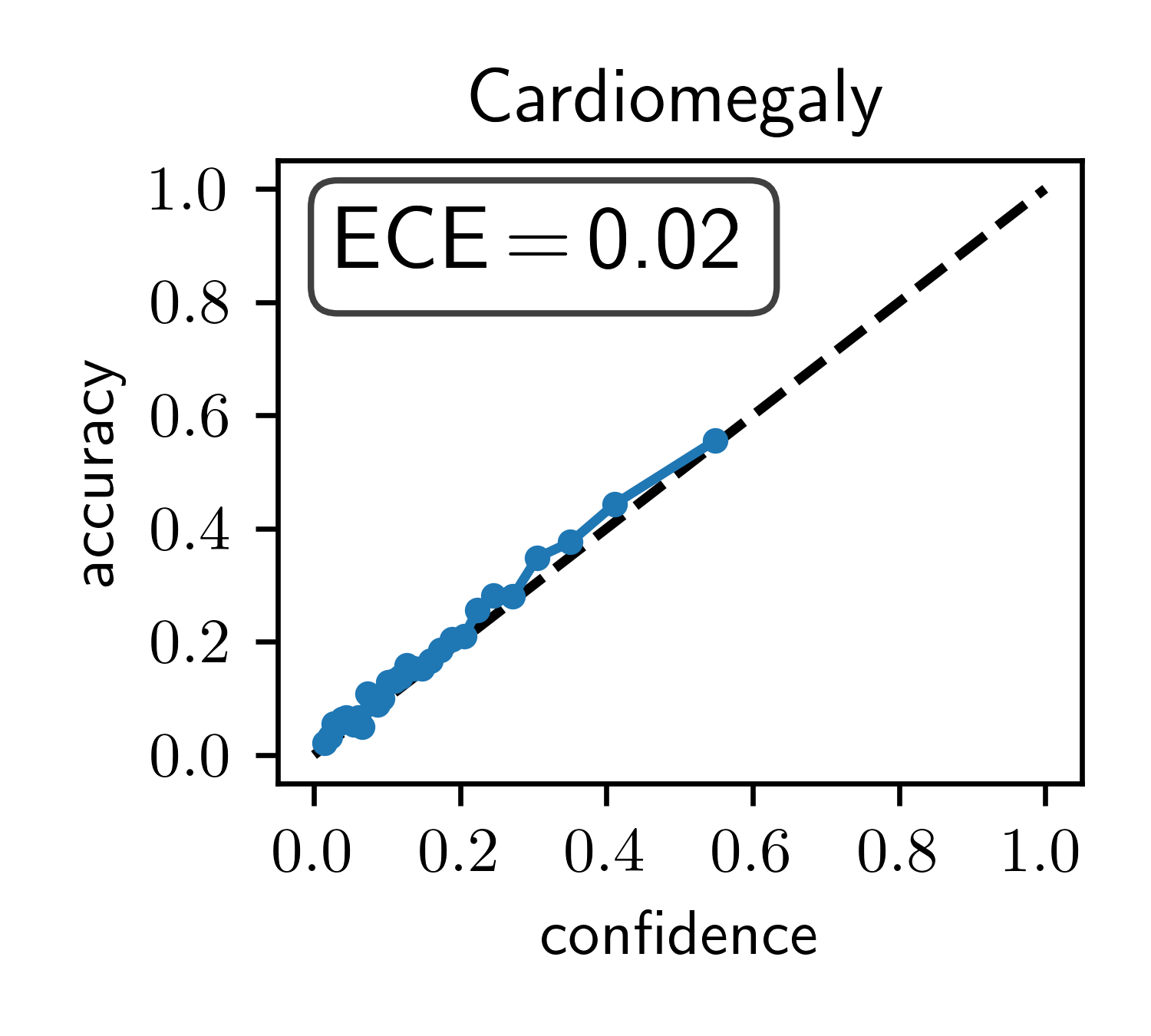}
     \includegraphics[width=0.19\textwidth]{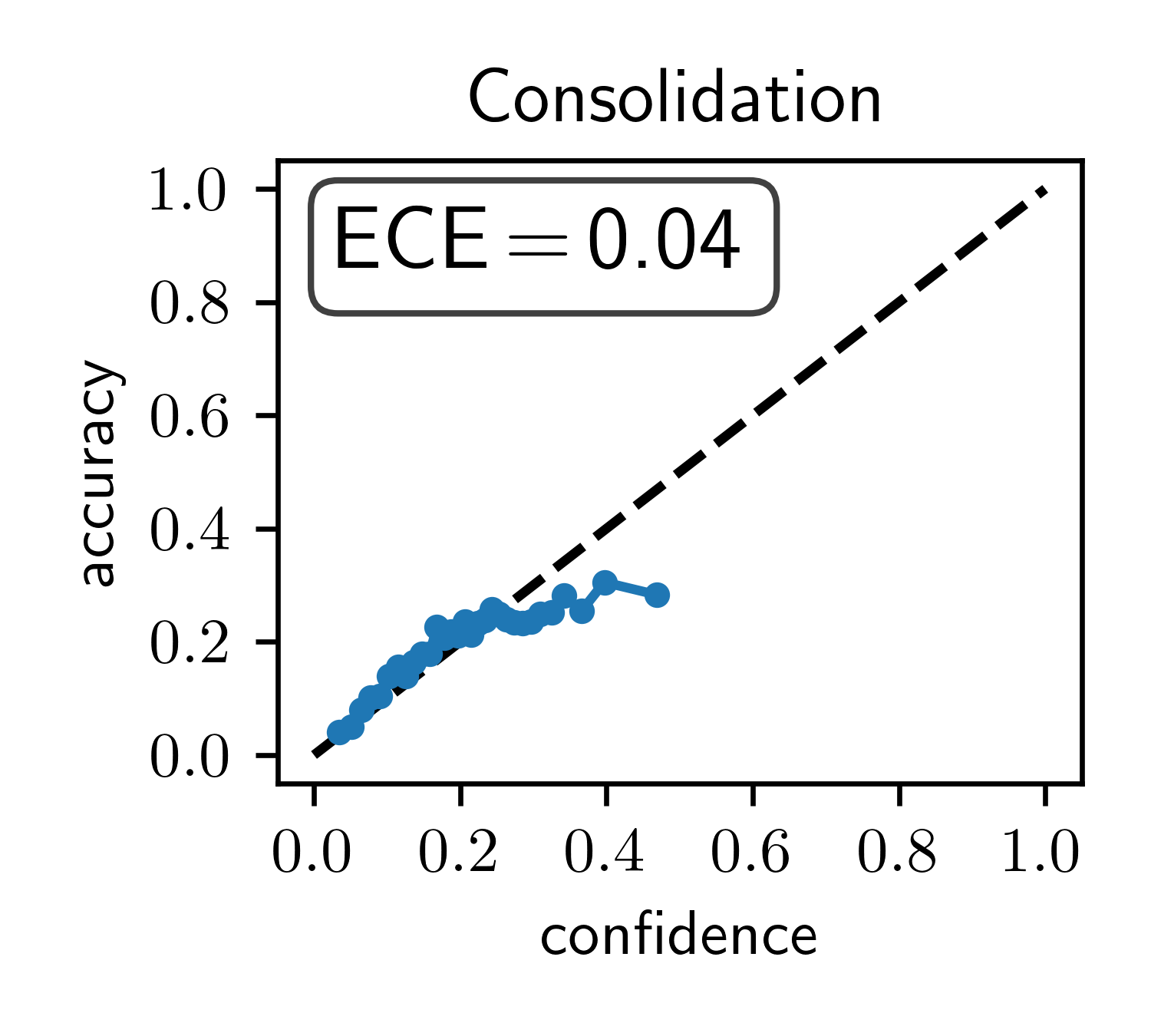}
     \includegraphics[width=0.19\textwidth]{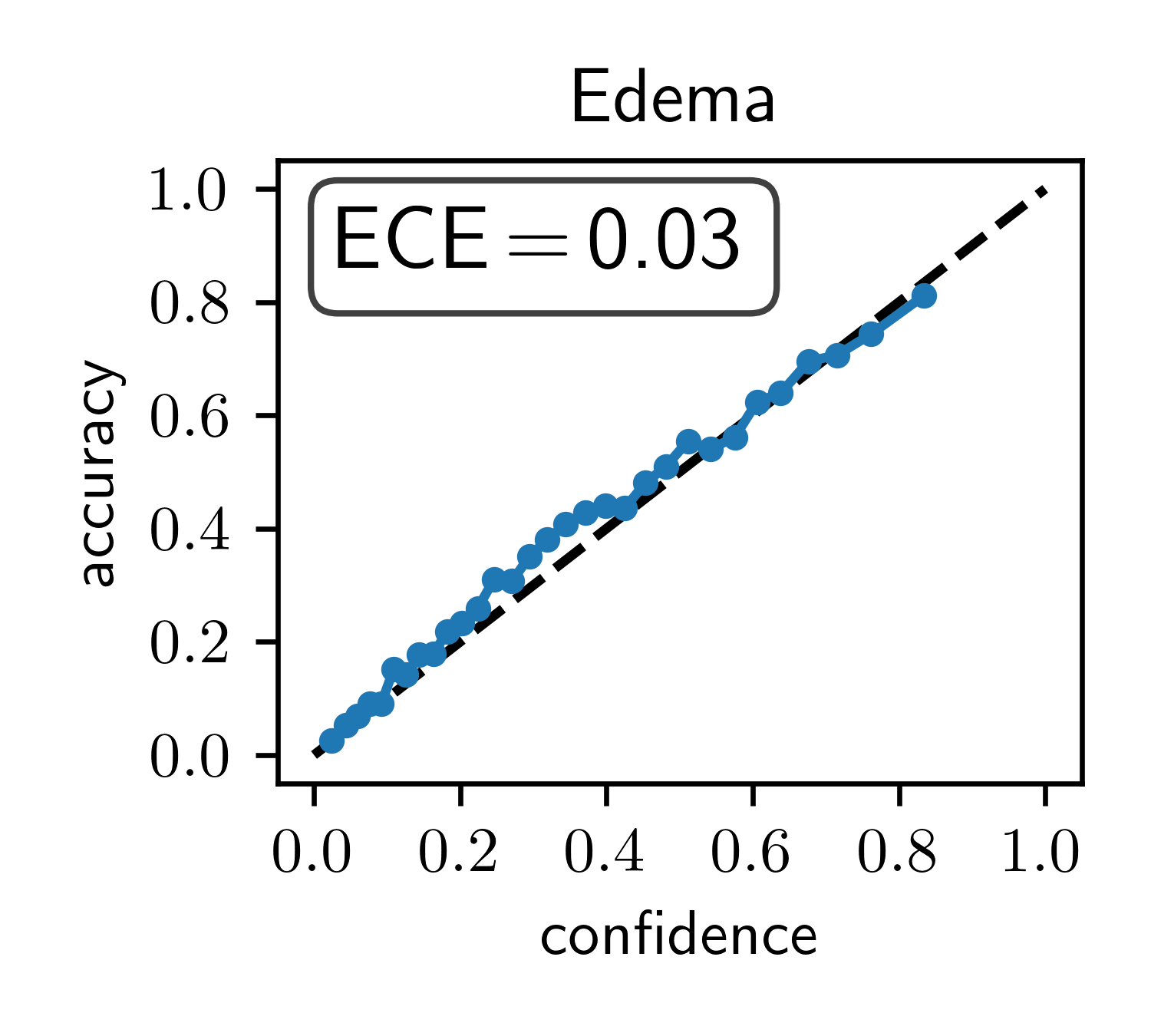}
     \includegraphics[width=0.19\textwidth]{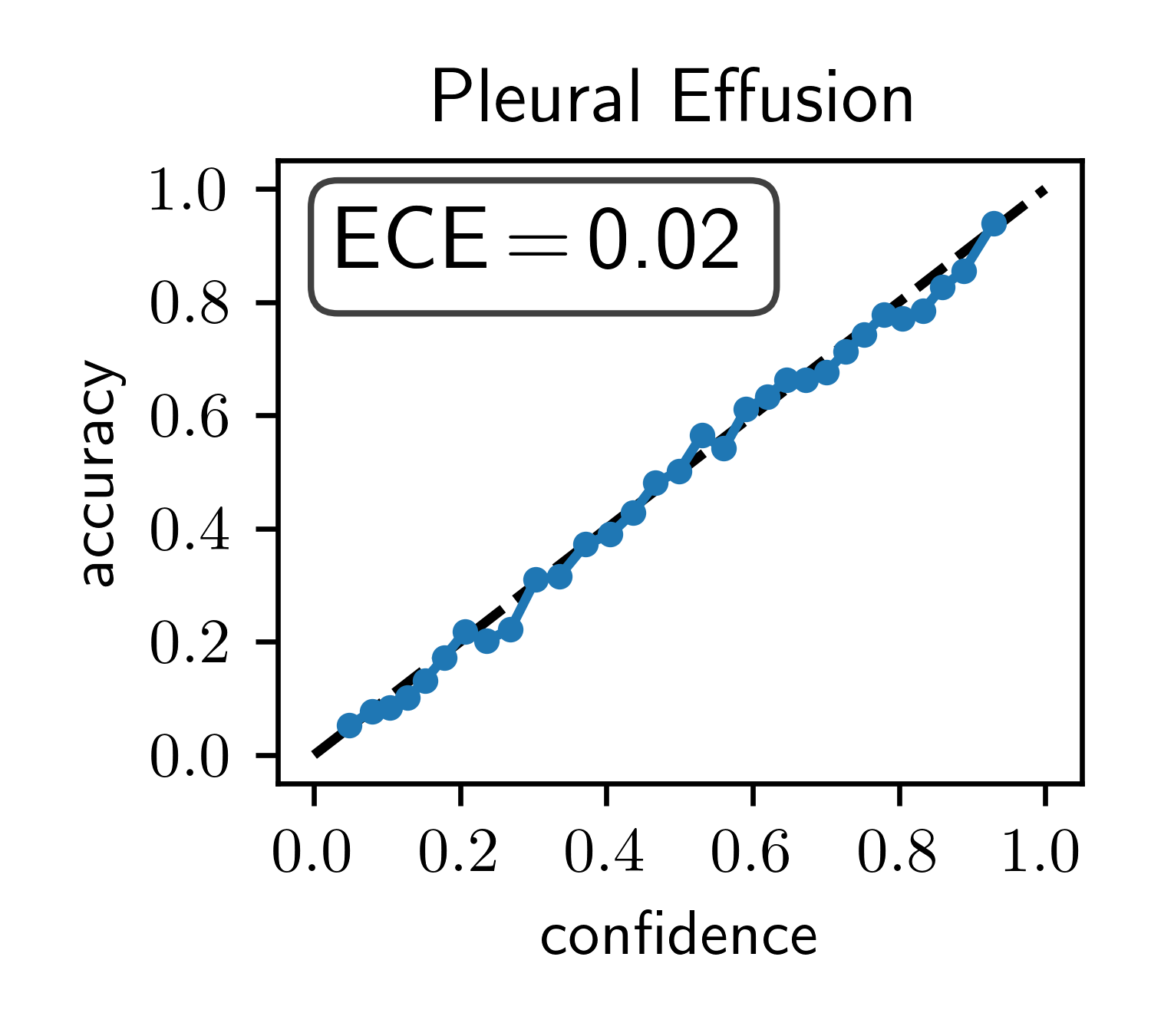}
    \caption*{BiT-50x1 with 5000 samples}
    
     \includegraphics[width=0.19\linewidth]{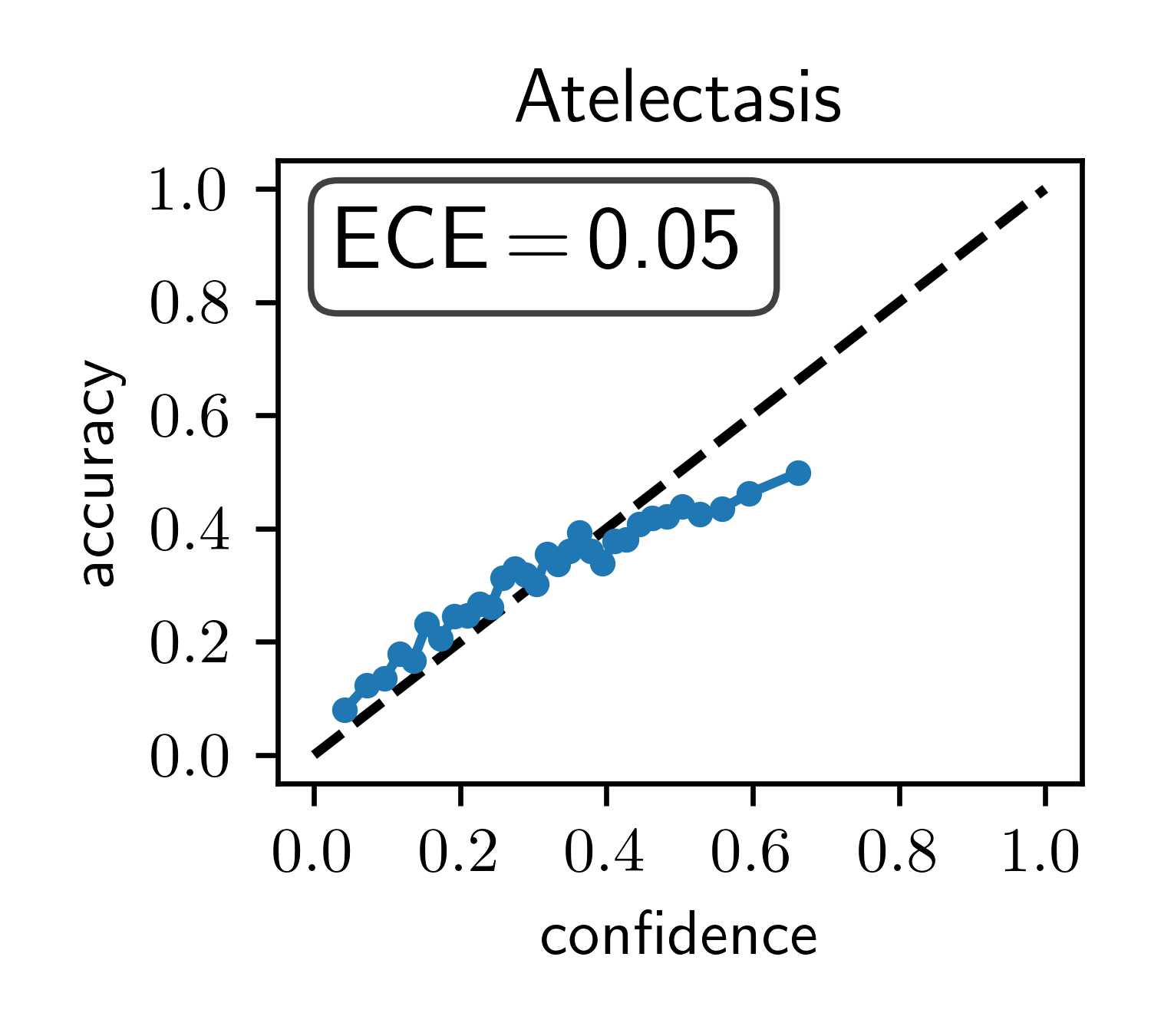}
     \includegraphics[width=0.19\textwidth]{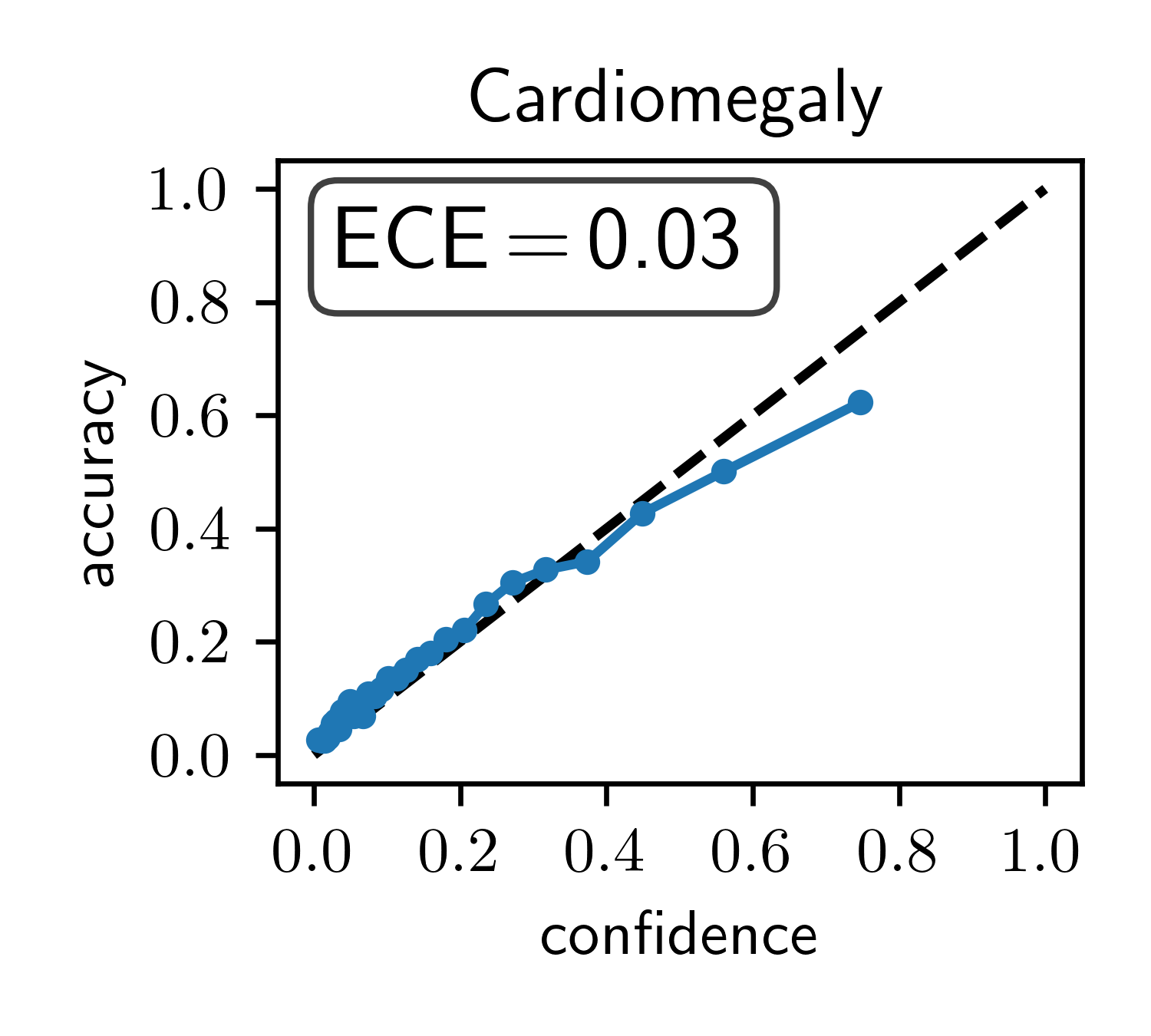}
     \includegraphics[width=0.19\textwidth]{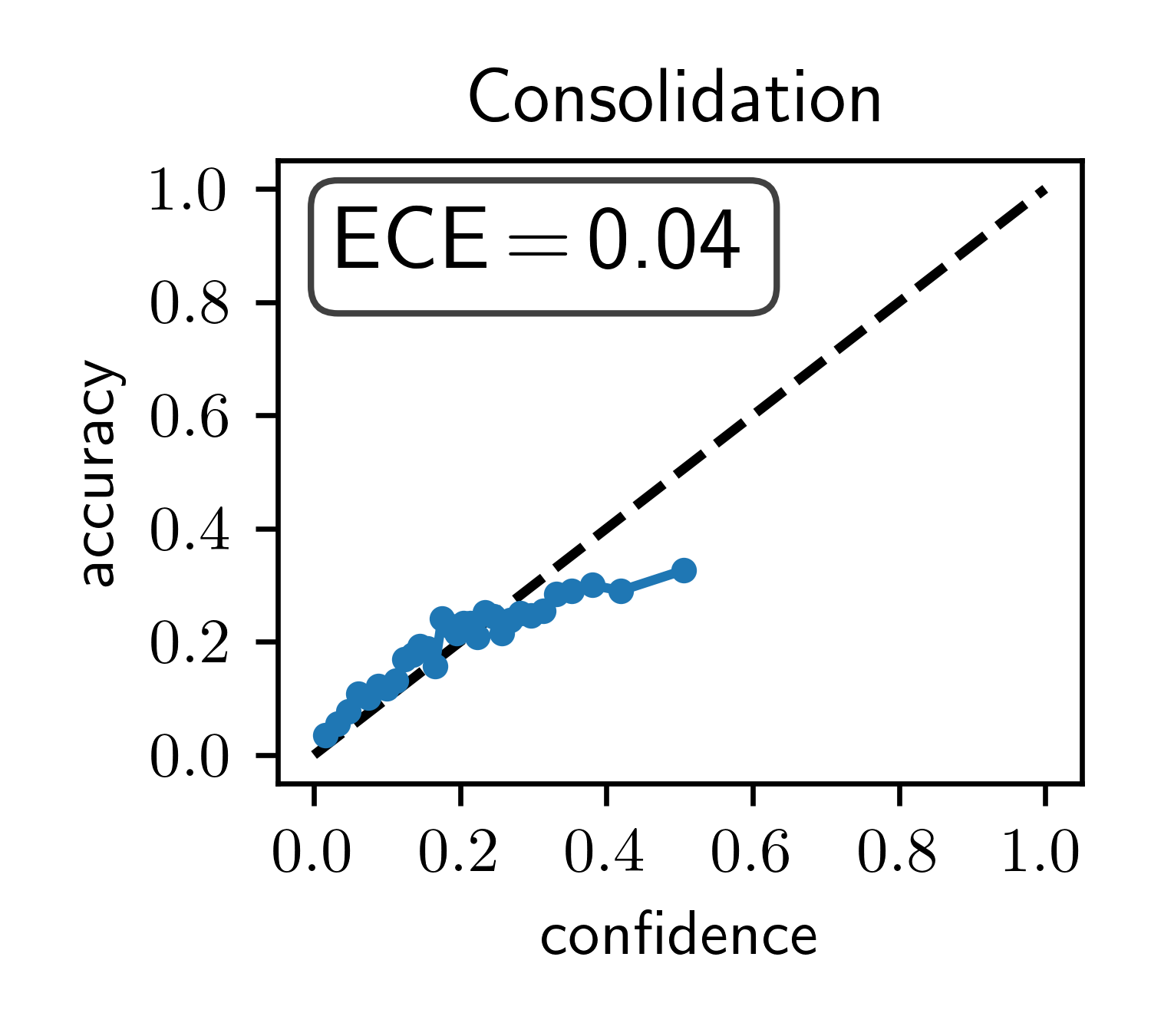}
     \includegraphics[width=0.19\textwidth]{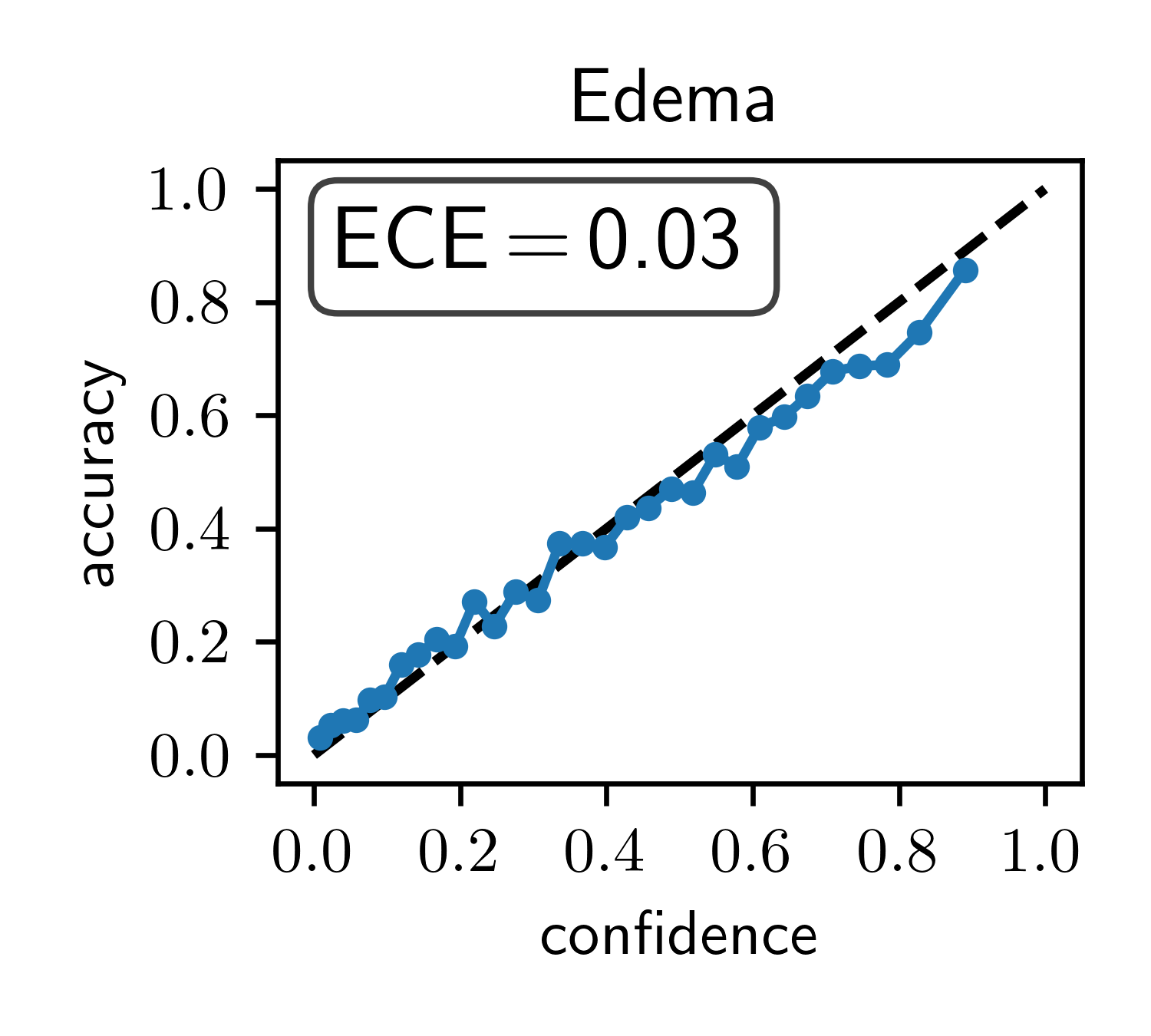}
     \includegraphics[width=0.19\textwidth]{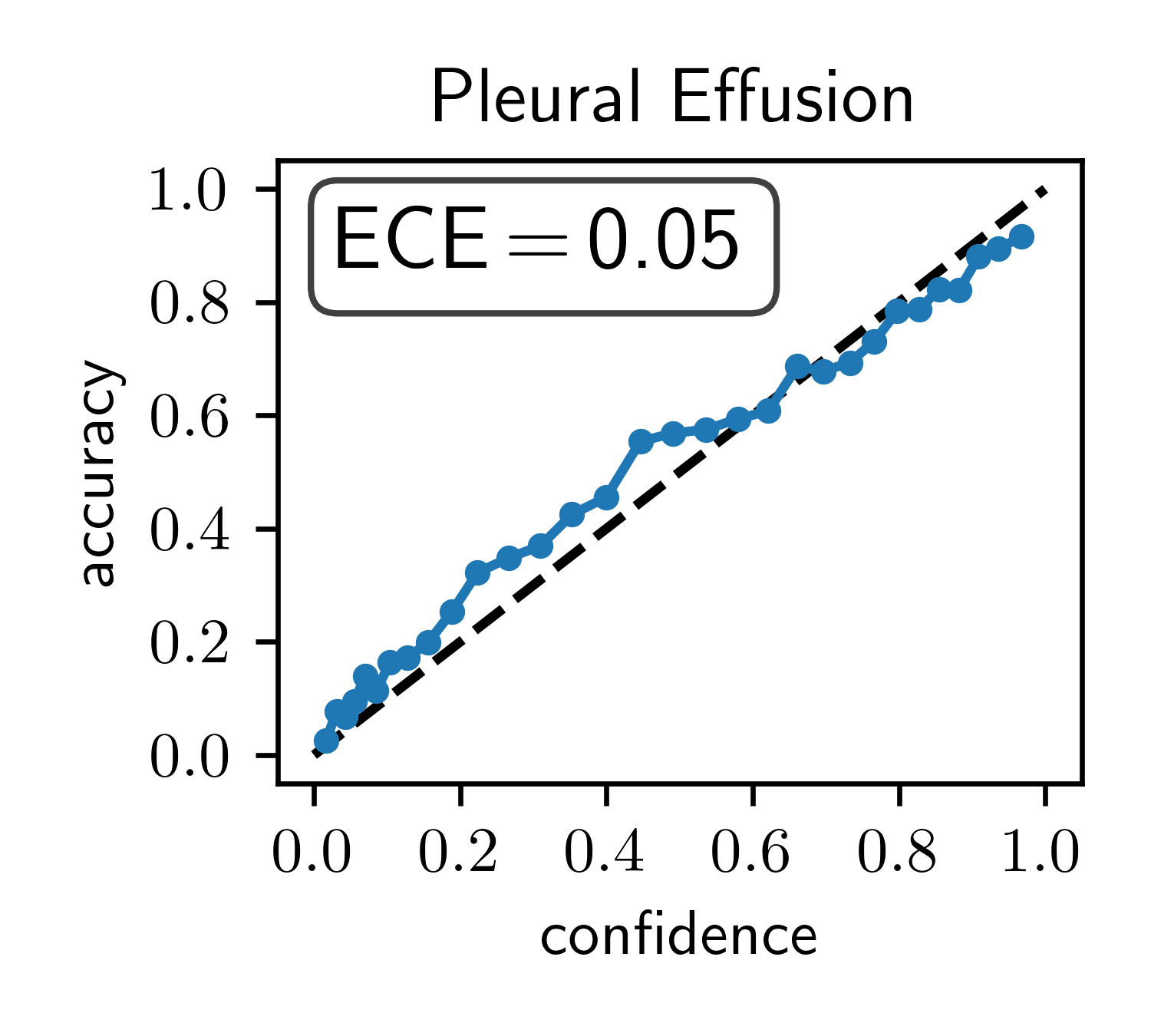}
    \caption*{BiT-101x3 with 5000 samples}
    
    \includegraphics[width=0.19\linewidth]{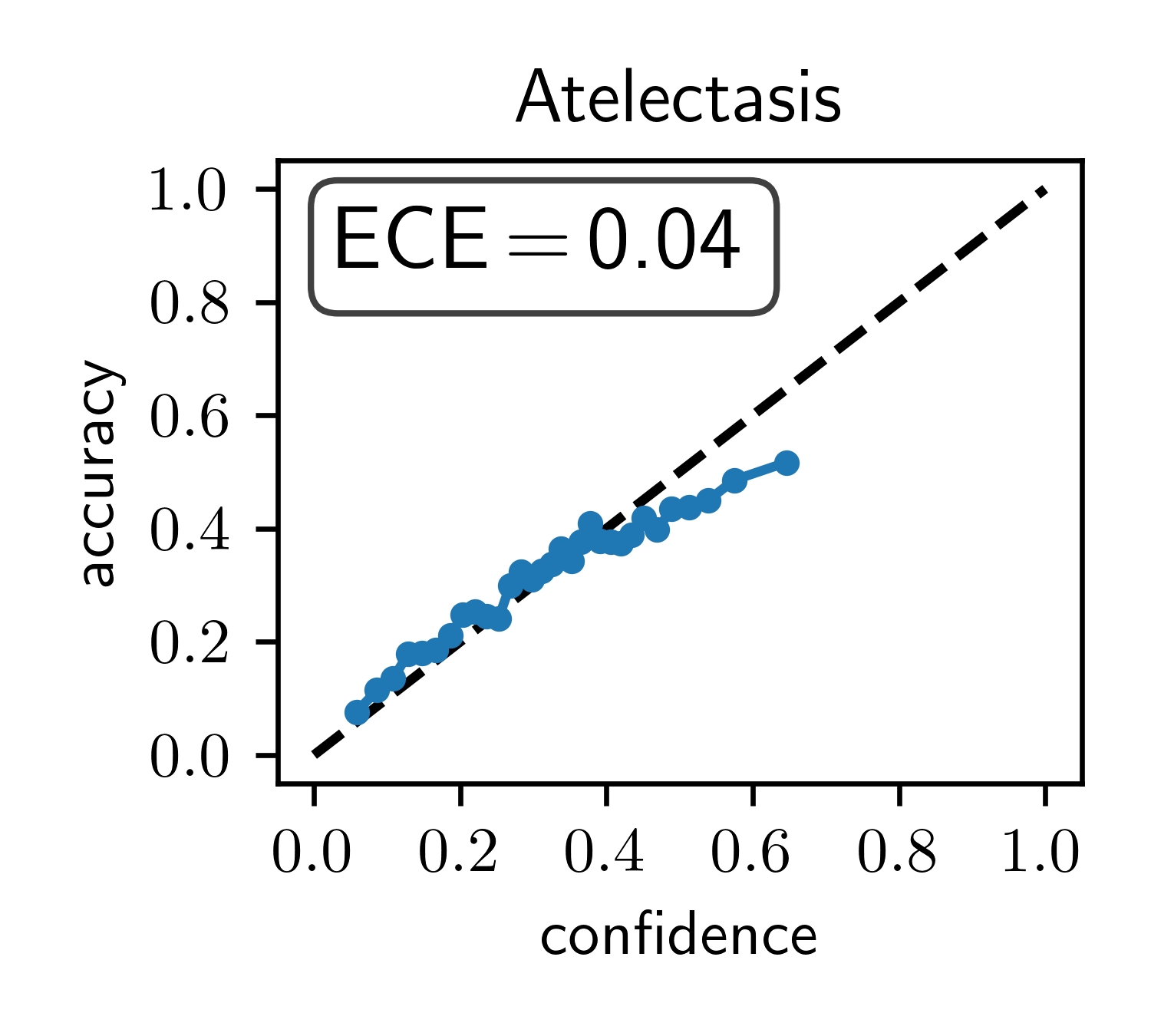}
     \includegraphics[width=0.19\textwidth]{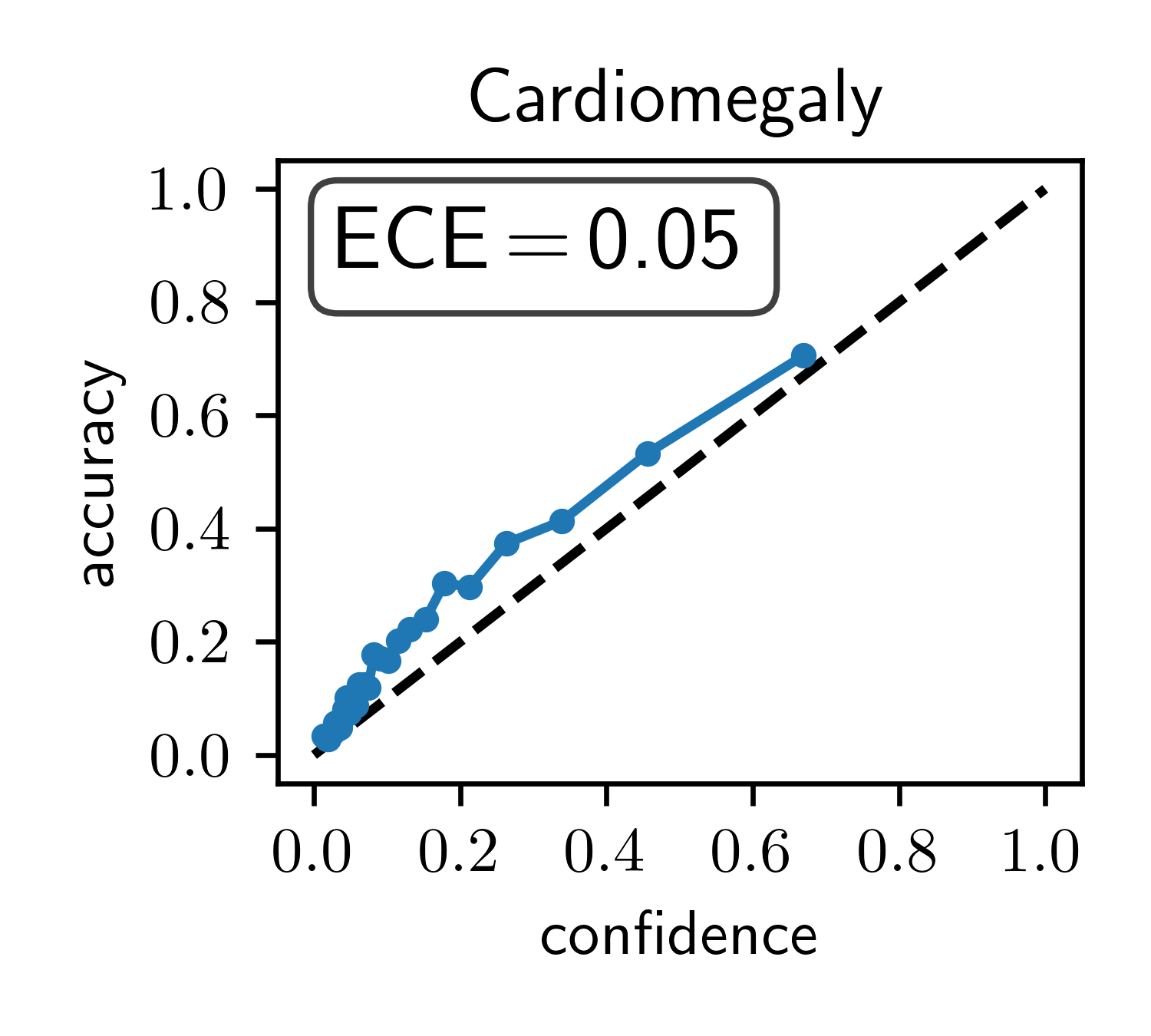}
     \includegraphics[width=0.19\textwidth]{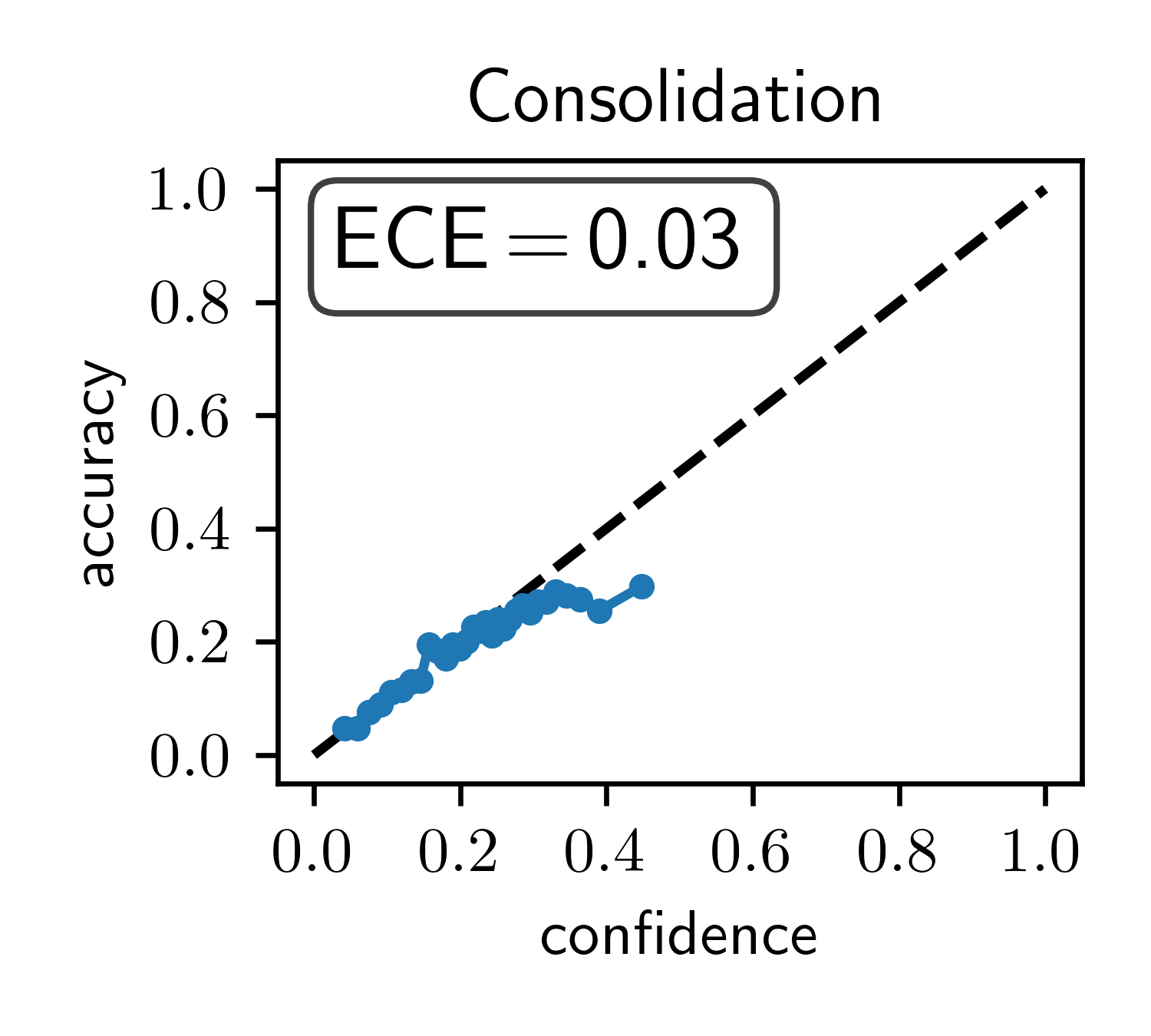}
     \includegraphics[width=0.19\textwidth]{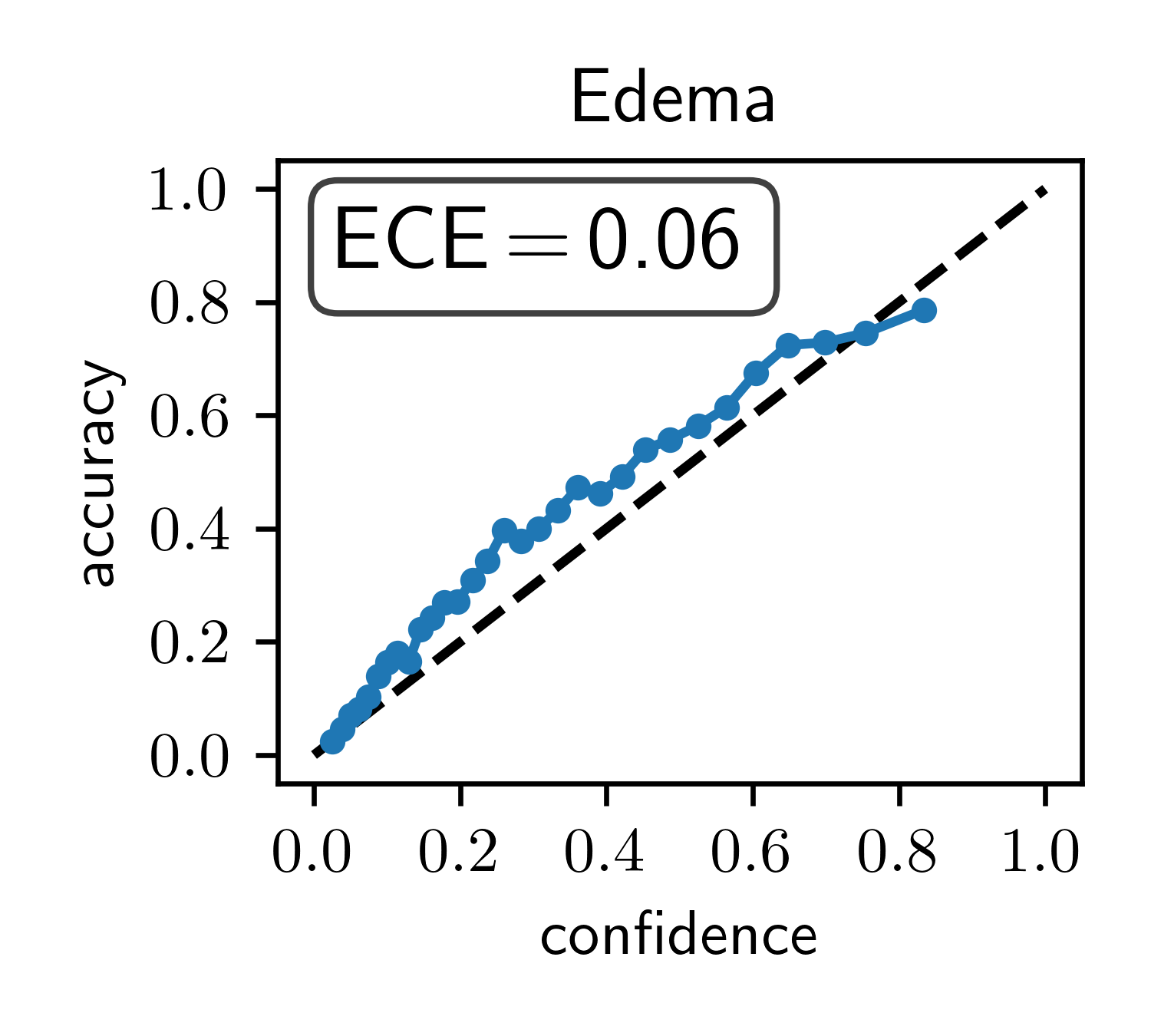}
     \includegraphics[width=0.19\textwidth]{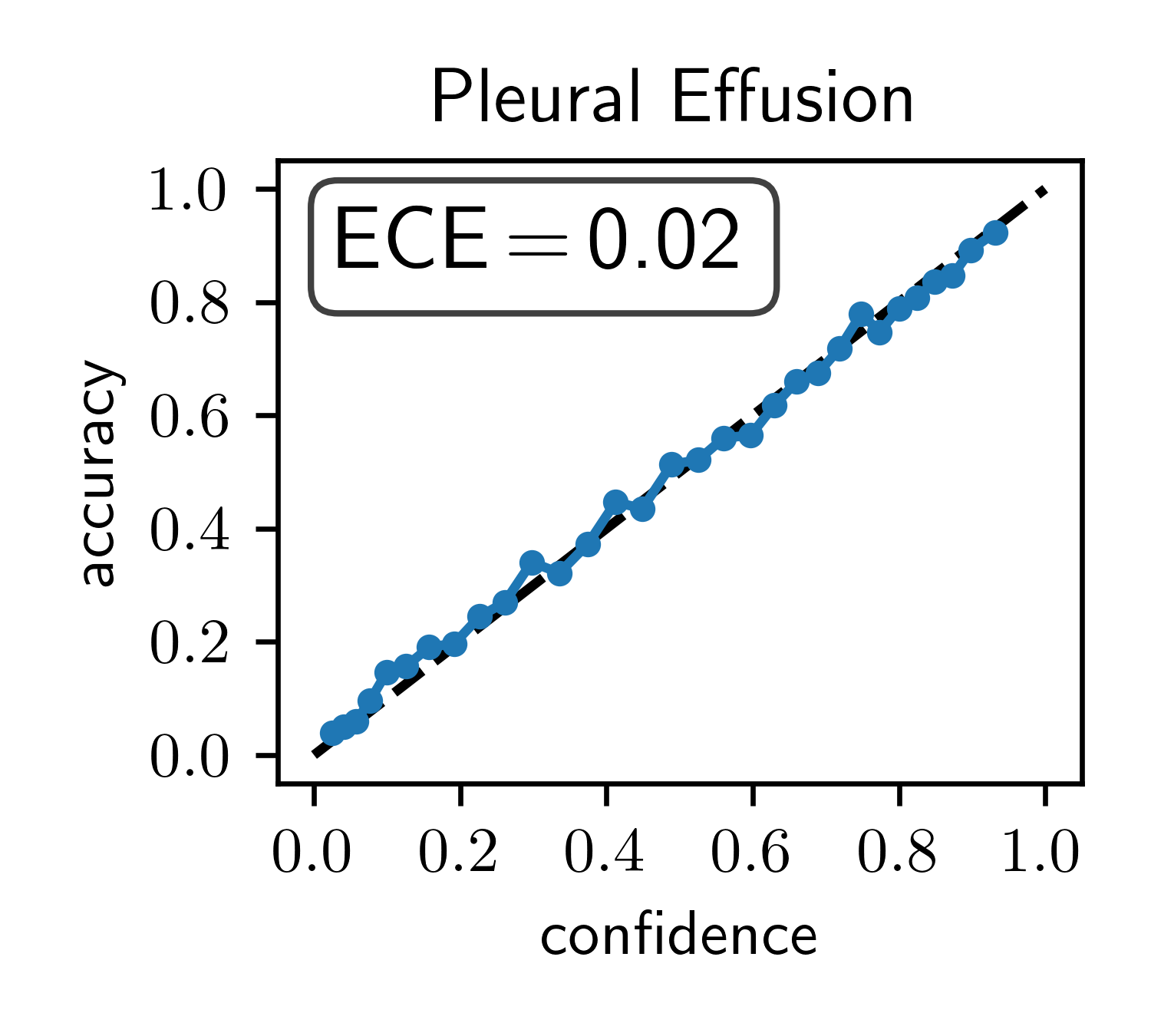}
    \caption*{ViT-M/P16 with 5000 samples}
    \caption{Calibration plots for model comparison on resplit test set. A perfectly calibrated model shows perfect correlation between confidence and class accuracy (dotted line). Average number of appearance in CheXpert5000 train sets: Atelectasis: 1623 (32.46 \%), Cardiomegaly: 815 (16.30\%), Consolidation: 1029 (20.58\%), Edema: 1775 (35.50\%), and Pleural Effusion: 2357 (47.14\%). 
    It can be seen that 'pleural effusion' and 'edema' have high confidence predictions while there are no model predictions with high confidence for 'atelectasis' or 'consolidation'.
    Overall, the class distribution is not a reliable indicator for accuracy/confidence see, e.\,g., 'cardiomegaly'.
The top row shows BiT-50x1 trained on 89944 training samples. 
The poor prediction quality (no predictions with high confidence) of the models is therefore not predominantly caused by the low data regime.
Even though some classes show an overall low accuracy, there is a high correlation between model predictions and accuracy for all models and all classes in both regimes. }
    
    \label{fig:calibration-plots}
\end{figure}

We provide the class-wise error metrics AUC and AUPRC in Tables~\ref{tab:c-AUC} and \ref{tab:c-AUPRC} as mentioned in Section \ref{sec:results}. 
We further provide reliability plots (calibration plots) for a selection of models, see Figure \ref{fig:calibration-plots}. 
The reliability plots are computed from 31 bins where each bin has an equal amount of samples.  Confidence is computed from the average model prediction per bin.
Additionally to the calibration of the models these plots show the distribution of model predictions for the five classes. An analysis is given in the caption.

\setlength{\tabcolsep}{5pt}

\begin{table}[]
\centering
\caption{Class-wise AUC on official CheXpert validation set. Average number of appearance in CheXpert5000 train sets: Atelectasis: 1623 (32.46 \%), Cardiomegaly: 815 (16.30\%), Consolidation: 1029 (20.58\%), Edema: 1775 (35.50\%), and Pleural Effusion: 2357 (47.14\%).}
\label{tab:c-AUC}
\begin{tabular}{@{}lcccccc@{}}
\toprule
\textbf{Model} &
  \textbf{mean} &
  \textbf{Atelectasis} &
  \textbf{Cardiom.} &
  \textbf{Consolid.} &
  \textbf{Edema} &
  \textbf{\begin{tabular}[c]{@{}c@{}}Pleural\\ Effusion\end{tabular}} \\ \midrule
\textbf{ResNet50}   & 0.7949  & 0.7815 & 0.6973 & 0.8390 & 0.8467  & 0.8102\\
\textbf{BiT-50x1}  & 0.\textbf{8380} & \textbf{0.7885}  & 0.8044 & \textbf{0.8568} & 0.8834  & 0.8568          \\
\textbf{BiT-50x3}  & 0.8359 & 0.7709  & 0.8056 & 0.8228 & 0.8949 & \textbf{ 0.8853}          \\
\textbf{BiT-101-3} & 0.8342 & 0.7671 & 0.8110 & 0.8166 & \textbf{0.8995} & 0.8768  \\
\textbf{\begin{tabular}[c]{@{}l@{}}BiT-50x1\\ + MixUp\end{tabular}} &
  0.8313 & 0.7725 & 0.8038 & 0.8445 & 0.8840 & 0.8516 \\
\textbf{ViT-B P16}  & 0.8299 & 0.7746 & \textbf{0.8146} & 0.8197 & 0.8735          & 0.8672 \\ \bottomrule
\end{tabular}
\end{table}

\begin{table}[]
\centering
\caption{Class-wise AUPRC on official CheXpert validation set. Average number of appearance in CheXpert5000 train sets: Atelectasis: 1623 (32.46 \%), Cardiomegaly: 815 (16.30\%), Consolidation: 1029 (20.58\%), Edema: 1775 (35.50\%), and Pleural Effusion: 2357 (47.14\%).}
\label{tab:c-AUPRC}
\begin{tabular}{@{}lcccccc@{}}
\toprule
\textbf{Model} &
  \textbf{mean} &
  \textbf{Atelectasis} &
  \textbf{Cardiom.} &
  \textbf{Consolid.} &
  \textbf{Edema} &
  \textbf{\begin{tabular}[c]{@{}c@{}}Pleural\\ Effusion\end{tabular}} \\ \midrule
\textbf{ResNet50}   & 0.5769 & \textbf{0.6585} & 0.4881 & 0.\textbf{4328} & 0.6396 & 0.6658 \\
\textbf{BiT-50x1}  & 0.6196 & 0.6238 & 0.6687 & 0.4162 & 0.7046 & 0.6845 \\
\textbf{BiT-50x3}  & 0.6218 & 0.5797 & 0.6512 & 0.3610 & 0.7455 & 0.\textbf{7717}\\
\textbf{BiT-101-3} & \textbf{0.6275} & 0.5846 & \textbf{0.6796} & 0.3681 & \textbf{0.7484} & 0.7567 \\
\textbf{\begin{tabular}[c]{@{}l@{}}BiT-50x1\\ + MixUp\end{tabular}} &
  0.6113 & 0.5959 & 0.6791 & 0.3966 & 0.7033 & 0.6815 \\
\textbf{ViT-B P16}  & 0.6269 & 0.6025 & 0.6775 & 0.4030 & 0.6941 & 0.7574 \\ \bottomrule
\end{tabular}
\end{table}








\end{document}